\documentclass{article}

     \PassOptionsToPackage{numbers, compress}{natbib}
\usepackage[preprint]{neurips_2026}

\usepackage[utf8]{inputenc} % allow utf-8 input
\usepackage[T1]{fontenc}    % use 8-bit T1 fonts
\usepackage{hyperref}       % hyperlinks
\usepackage{url}            % simple URL typesetting
\usepackage{booktabs}       % professional-quality tables
\usepackage{amsfonts}       % blackboard math symbols
\usepackage{nicefrac}       % compact symbols for 1/2, etc.
\usepackage{microtype}      % microtypography
\usepackage{xcolor}         % colors

\title{Vision-driven Preference Synthesis for \\ Mitigating Hallucinations in VLMs}

\author{
  Yunhun Nam \qquad Jongheon Jeong\\
  Korea University \\
  \texttt{\{yh0326, jonghj\}@korea.ac.kr} \\
}

\usepackage{amsmath,amsfonts,bm}

\def\1{\bm{1}}

\def\vtheta{{\bm{\theta}}}

\def\vc{{\bm{c}}}

\def\vm{{\bm{m}}}

\def\vq{{\bm{q}}}

\def\vx{{\bm{x}}}
\def\vy{{\bm{y}}}

\DeclareMathAlphabet{\mathsfit}{\encodingdefault}{\sfdefault}{m}{sl}
\SetMathAlphabet{\mathsfit}{bold}{\encodingdefault}{\sfdefault}{bx}{n}

\newcommand{\ie}{\textit{i}.\textit{e}.}

\usepackage{mwe}
\usepackage{float}
\usepackage{wrapfig}
\usepackage{adjustbox}
\usepackage{multirow} 
\usepackage[table]{xcolor}
\usepackage{wrapfig}
\usepackage{subcaption}
\usepackage[table]{xcolor}
\usepackage{amssymb}

\usepackage{amsmath, amssymb}

\usepackage{algorithm}
\usepackage{algpseudocode}

\algrenewcommand\algorithmicrequire{\textbf{Input:}}
\algrenewcommand\algorithmicensure{\textbf{Output:}}
\algrenewcommand\algorithmiccomment[1]{\hfill{\footnotesize\color{gray}// #1}}

\definecolor{vanillagray}{RGB}{242,242,242}
\definecolor{ourscyan}{RGB}{225,245,250}

\definecolor{pinegreen}{rgb}{0.0, 0.47, 0.44}
\definecolor{cornellred}{rgb}{0.7, 0.11, 0.11}
\definecolor{cadmiumgreen}{rgb}{0.0, 0.42, 0.24}
\definecolor{spirodiscoball}{rgb}{0.06, 0.75, 0.99}
\definecolor{navyblue}{rgb}{0,0.4,0.8}
\definecolor{Red7}{rgb}{0.941, 0.243, 0.243}
\definecolor{Green7}{RGB}{55, 178, 77}
\definecolor{Blue9}{rgb}{0.098,0.3,0.9}
\definecolor{darkred}{rgb}{0.55, 0.0, 0.0}
\definecolor{mygreen}{rgb}{0, 0.6823, 0.7215}

\usepackage{pifont}

\hypersetup{
  linkcolor = cornellred,
  citecolor  = cadmiumgreen,
  colorlinks = true,
  urlcolor = Blue9
}

\renewcommand{\paragraph}[1]{\vspace{1mm}\noindent\textbf{#1.}\,\,}

\newcommand{\methodname}{ViPSy\xspace}

\begin{document}

\maketitle

\begin{abstract}
\label{sec:abstract}
Vision-Language Models (VLMs) have shown strong performance in visual understanding, yet they still suffer from hallucinations, generating content that is not grounded in the image. Preference alignment is a promising approach to improve visual faithfulness, but its success depends heavily on how preference pairs are constructed. Existing methods exhibit two key limitations; {(a) intervention-based methods often introduce significant deviation from the policy distribution, and (b) sampling-based methods often underuse visual information during the construction.}
{In this paper,} we propose \methodname{} (\textbf{Vi}sion-driven \textbf{P}reference \textbf{Sy}nthesis), a framework for constructing preference data that are both policy-aligned and visually grounded. Our framework consists of two stages; in the first stage, \methodname{} derives a visual cue from recurring object-level content across semantically aligned image variants, so preference construction can rely on visual information rather than language priors. In the second stage, \methodname{} conditions the policy's own rollouts on this cue, allowing candidates to be guided by visually grounded content while staying close to the policy's response distribution. The resulting candidates remain close to the policy's response distribution while better leveraging visual information from the image. Experiments show that the resulting VLM, preference-aligned with \methodname{}-constructed preference pairs, achieves a new state-of-the-art in hallucination mitigation. Compared with the {previous} state-of-the-art method, it reduces hallucination rates on AMBER and Object HalBench by 35.7\% and 24.5\%, respectively. The resulting model further improves on general visual grounding benchmarks, e.g., MMStar, MMVP, and CV-Bench, while also yielding gains in semantic segmentation and ImageNet linear probing, underscoring the effectiveness of our framework in enhancing the model's visual capabilities. Code is available at \url{https://github.com/yunpal/ViPSy}.
\end{abstract}
\section{Introduction}

Recent progress in Vision-Language Models (VLMs) \cite{liu2024improved, alayrac2022flamingo, dai2023instructblip, bai2025qwen3, 2023GPT4VisionSC} has led to notable improvements in visual understanding. By combining pretrained vision encoders \cite{radford2021learning, zhai2023sigmoid} with Large Language Models (LLMs) \cite{vicuna2023, hui2024qwen2}, VLMs can jointly process images and text, supporting vision-language applications across diverse domains, including education \cite{lu2023mathvista, lee2024llava_edu}, medical applications \cite{li2023llava_med, luo2023biomedgpt, moor2023med}, industrial inspection \cite{jeong2023winclip, gu2024anomalygpt, jiang2024mmad}, and robotics \cite{xiong2024aic, liu2024enhancing, li2024manipllm}. Despite these capabilities, VLMs still suffer from a fundamental limitation known as \emph{hallucination} \cite{gunjal2024detecting, guan2024hallusionbench}, \ie, generating content inconsistent with the given image, often introducing objects that do not actually exist. 
This limitation raises concerns about the reliability of VLMs in real-world applications, making hallucinations a critical challenge to address.

\begin{figure}[t]
    \centering
    \includegraphics[width=\linewidth]{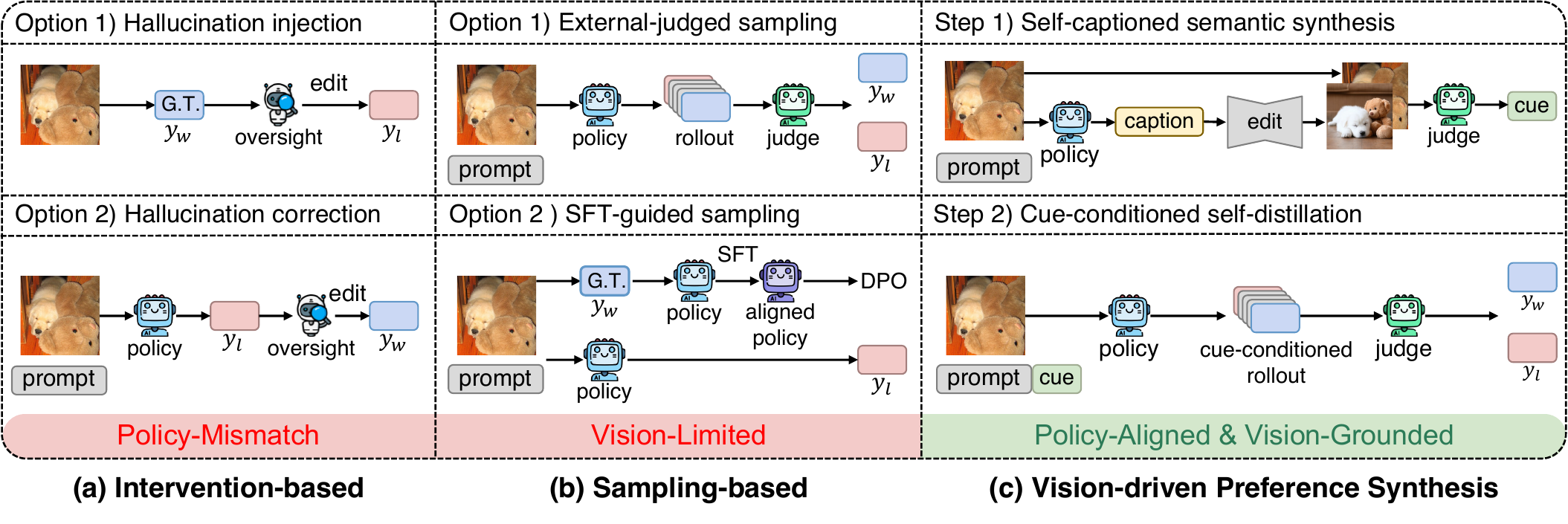}
    \caption{
    Comparison of preference construction strategies for hallucination mitigation in VLMs.
    (a) Intervention-based methods can create preference pairs that deviate from the policy distribution.
    (b) Sampling-based methods reduce this mismatch, but underuse visual information.
    (c) \methodname{} addresses both issues through self-captioned semantic synthesis for consistent cue extraction and cue-conditioned self-distillation for preference synthesis.
    }
    \label{fig:concept}
\end{figure}

Hallucination in VLMs often arises from imbalances between the language and vision modalities, where strong language priors make models rely on text instead of the image \cite{wu2022overcoming, han2022visual, zhibo2023overcoming, lee2024volcano, leng2024mitigating}. Hallucination is also associated with statistical biases in pre-trained data \cite{agrawal2016analyzing, goyal2017making, agarwal2020towards, huang2025shield}, which can cause models to favor frequent but visually unsupported associations. To address these issues, {preference-based alignment} \cite{zhao2023beyond, peng2025mitigating, yang2025opadpo} has emerged as a promising direction for improving the visual faithfulness of VLMs. In particular, hallucination mitigation can be naturally formulated as a preference optimization, where hallucination-free responses are treated as preferred outputs over hallucinated ones under the same image and prompt. For hallucination mitigation in VLMs, Direct Preference Optimization (DPO) \cite{rafailov2023direct} has become a practical choice for preference alignment, as it directly optimizes the model from preference pairs without explicit reward modeling or online rollouts. By avoiding these requirements, DPO sidesteps the challenges of reinforcement learning (RL), which updates a policy to maximize preference-derived rewards \cite{ziegler2019fine, stiennon2020learning, ouyang2022training, bai2022training}. Specifically, reliable image-grounded rewards are difficult to define due to vision-language misalignment \cite{sun2024aligning}, and RL-based methods further require reward modeling and repeated online rollouts \cite{rafailov2023direct, casper2023open, sidahmed2024parameter, li2025provably}.

Nevertheless, the effectiveness of DPO largely depends on the quality of the constructed preference pairs \cite{ivison2024unpacking, pan2025matters}. As shown in Figure~\ref{fig:concept}, existing methods for hallucination mitigation differ mainly in how they construct such pairs and can be grouped into (a) intervention-based and (b) sampling-based methods. Intervention-based methods include hallucination injection methods \cite{zhou2024aligning, sarkar2024halva}, which create rejected responses by inserting hallucinated content into correct responses, and hallucination correction methods \cite{sun2024aligning, zhao2023beyond, xiao2025detecting}, which revise model-generated responses. Although these methods can provide explicit hallucinated and hallucination-free pairs, their responses are partially determined by external intervention and thus deviate from the policy distribution. Sampling-based methods rather aim to reduce this distribution mismatch by making preference data closer to the policy \cite{yu2024rlaif, yang2025opadpo,peng2025mitigating}. External-judged sampling methods construct preference pairs from the policy's rollouts and use a judge to determine preferred and dispreferred responses \cite{yu2024rlaif, peng2025mitigating}. Supervised fine-tuning (SFT)-guided sampling first performs SFT on hallucination-free responses and then aligns the resulting model with pairs that favor SFT-aligned outputs over pre-SFT outputs \cite{yang2025opadpo}. Although policy-aligned preference data can improve the effectiveness of preference optimization \cite{agarwal2024policy, yang2024self, zhao2026self}, existing methods still lack the use of visual information.

\paragraph{Contributions}
In this paper, we propose \methodname{} (\textbf{Vi}sion-driven \textbf{P}reference \textbf{Sy}nthesis), a framework for hallucination mitigation that constructs visually grounded preference pairs while keeping them close to the policy's response distribution. The framework addresses two challenges in preference construction: limited use of visual information and distribution mismatch with the policy. First, we introduce \textit{self-captioned semantic synthesis} for consistent cue extraction. Using the policy's own self-captions as T2I prompts, we synthesize semantically aligned image variants. Comparing these variants with the original image reveals recurring object-level evidence, which is aggregated into a consistent visual cue. This cue provides explicit visual guidance for preference construction, reducing reliance on language priors. Second, we introduce \textit{cue-conditioned self-distillation} for preference synthesis. The consistent visual cue guides the policy's own rollouts, producing cue-conditioned self-generated candidate responses. We then select visually grounded and less grounded responses from these candidates. Thus, the cue encourages visual grounding, while the policy's own rollouts keep the resulting pairs close to its distribution.

Experimental results show that \methodname{} enables a new state-of-the-art in mitigating hallucinations of VLMs when used for DPO. Compared to the previous state-of-the-art results, our method significantly reduces AMBER CHAIR and Object HalBench response-level hallucination by 35.7\% and 24.5\%, respectively. Interestingly, we observe that our alignment scheme further improves general vision-language benchmarks that require visual grounding, with gains of 6.0\% on MMStar, 31.3\% on MMVP, and 4.0\% on CV-Bench. Moreover, the optimized vision encoder shows gains in patch-level segmentation probing and ImageNet linear probing, suggesting improvements in both response-level visual faithfulness and visual representations.

We summarize the key contributions of this work as follows:
\begin{itemize}
    \item We propose \methodname{}, a vision-driven preference synthesis framework that constructs visually grounded preference pairs while keeping them close to the policy's response distribution.

    \item We introduce a two-stage preference construction strategy: self-captioned semantic synthesis for extracting a consistent visual cue, and cue-conditioned self-distillation for generating policy-aligned candidate responses guided by the cue.

    \item We show that \methodname{} sets a new state-of-the-art in hallucination mitigation, improves general vision-language benchmarks requiring visual grounding, and further enhances the visual representations of the underlying vision encoder.
\end{itemize}

\section{Related work}
\label{sec:related}

\paragraph{Preference alignment}
Preference alignment aims to make model outputs better reflect human preferences. A representative approach is Reinforcement Learning from Human Feedback (RLHF) \cite{ouyang2022training, sun2024aligning}, which trains a reward model to judge rollout preferences and then optimizes the policy with PPO \cite{schulman2017proximal}. This framework has been used for aligning models to be helpful \cite{bai2022training}, while AI feedback can reduce reliance on human preference labeling \cite{bai2022constitutional, xiao2025detecting, yu2024rlaif}. Recent RL variants such as Group Relative Policy Optimization (GRPO) reduce policy-optimization cost with group-relative rewards \cite{shao2024deepseekmath}, but still require rollout-level reward evaluation, which is difficult to define for VLM hallucination mitigation under vision-language misalignment \cite{sun2024aligning}. DPO \cite{rafailov2023direct} instead learns from pre-constructed preferred and dispreferred response pairs, avoiding explicit reward modeling or online RL and shifting the key challenge to preference construction. Building on the same preference optimization perspective, related methods have explored alternative objectives, including sequence-level likelihood calibration \cite{zhao2023slic}, general theoretical objectives for learning from preferences \cite{azar2024general}, alignment from binary desirable or undesirable feedback \cite{ethayarajh2024kto}, and optimization without a reference model \cite{hong2024orpo, meng2024simpo}. Preference-based methods further improve visual faithfulness by modifying alignment objectives \cite{xie2024v, jiang2024modality, wang2024mdpo, liu2025mitigating}. These works provide diverse preference optimization objectives, while our work complements them by constructing visually grounded preference data for VLM hallucination mitigation.

\paragraph{Mitigating hallucination in VLMs}
Motivated by this shift from reward design to preference-data construction, hallucination mitigation in VLMs focuses on constructing visually grounded preference pairs, where preferred responses should be faithful to the image while dispreferred responses capture hallucinated or unsupported content. Preference-construction methods differ in how response pairs are obtained. Intervention-based methods construct pairs by modifying responses or visual inputs, including hallucination injection \cite{zhou2024aligning}, hallucination correction \cite{zhao2023beyond}, segment-level correction \cite{yu2024rlhf}, semantic-component manipulation \cite{he2024systematic}, phrase modification in ground-truth descriptions \cite{sarkar2024halva}, and distorted images \cite{pi2024strengthening}. In contrast, sampling-based methods derive pairs from policy-generated responses using external judges \cite{yu2024rlaif}, SFT-guided on-policy alignment \cite{yang2025opadpo}, detector-based verification~\cite{peng2025mitigating}, which relies on detector models \cite{liu2024grounding, cheng2024yolo}. Our work follows this sampling-based direction while further incorporating visual information through visual cues extracted from self-captioned semantic synthesis and cue-conditioned self-distillation for preference synthesis.

\paragraph{Policy-aligned self-generated data}
Sampling-based methods build on the observation that learning can benefit from
training data that is close to the policy distribution
\cite{tajwar2024preference, agarwal2024policy}. This perspective has motivated
a broad range of self-generation approaches, including self-generated
instructions \cite{wang2023self}, self-generated rewards
\cite{yuan2024self}, self-play preferences \cite{wu2024self}, synthetic preference data \cite{dong2024self}, and policy-aligned self-distillation \cite{yang2024self,zhao2026self}. However, these methods do
not explicitly account for visual inputs during self-generation. When an image
is provided, the resulting supervision can therefore be driven by language
priors rather than the visual information in the image. \methodname addresses
this limitation by conditioning the policy's own rollouts on consistent visual
cues, keeping the generated data close to the policy distribution while making
its construction explicitly vision-grounded.

\begin{figure}[t]
    \centering
    \includegraphics[width=\linewidth]{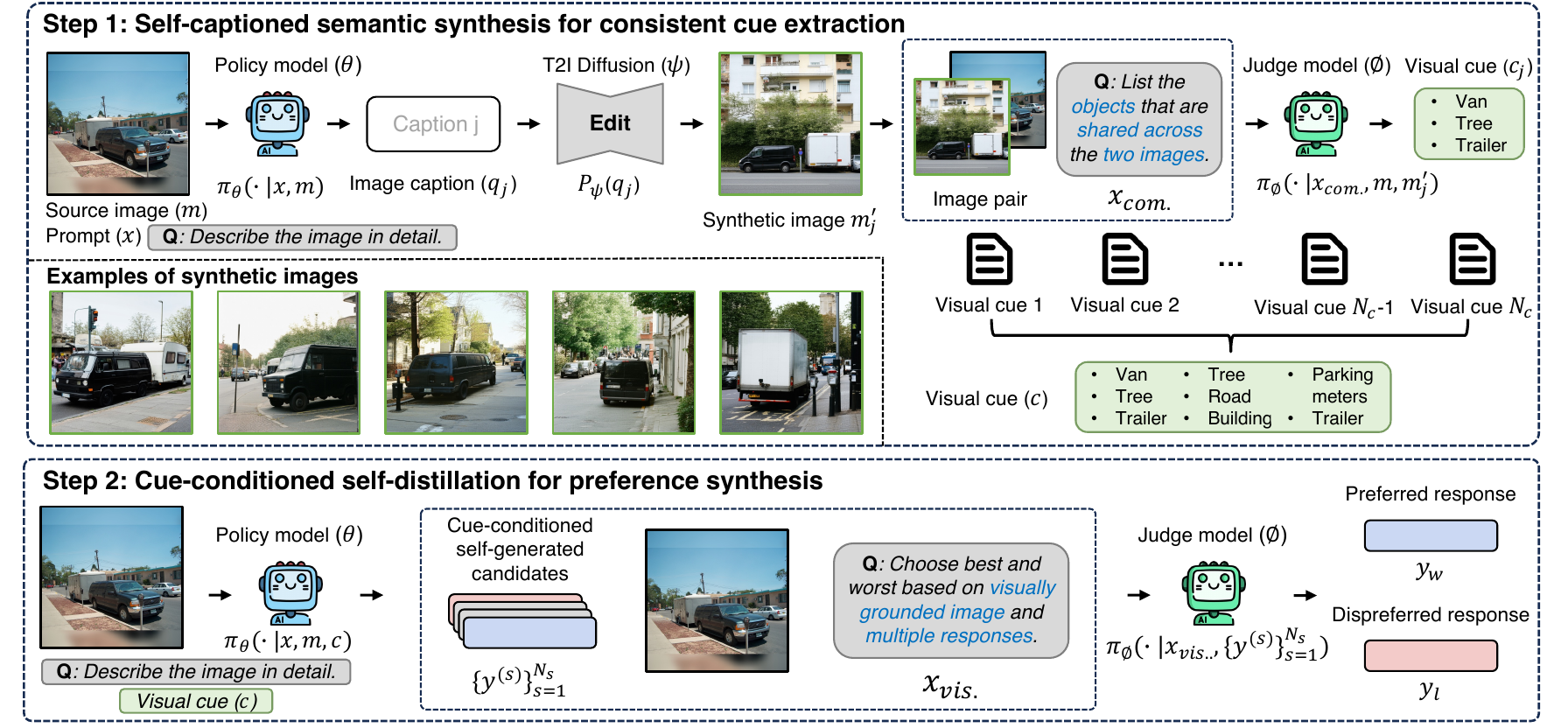}
    \caption{
    \textbf{Overview of \methodname{}.} 
    \methodname{} constructs visually grounded and policy-aligned preference pairs through two stages.
    (1) \textit{Self-captioned semantic synthesis for consistent cue extraction} uses the policy's own self-captions as T2I prompts to generate semantically aligned synthetic variants. Each variant is compared with the original image, and recurring object-level evidence across comparisons is aggregated into a consistent visual cue.
    (2) \textit{Cue-conditioned self-distillation for preference synthesis} uses the consistent visual cue to guide the policy's own rollouts, producing cue-conditioned self-generated candidate responses. A judge model then selects preferred and dispreferred responses according to their visual grounding with respect to the original image.
    }
    \label{fig:overview}
\end{figure}

\section{\methodname{}: Vision-driven Preference Synthesis}
\label{method}

We consider a vision-language model that generates a response $\vy$ conditioned on a prompt $\vx$ and an input image $\vm$, given a policy $\pi_\vtheta$ parameterized by $\vtheta$:

\begin{equation}
\pi_\vtheta(\vy \mid \vx, \vm)
= \prod_{t=1}^{T} \pi_\vtheta(\vy_t \mid \vx, \vm, \vy_{<t}).
\label{eq:policy}
\end{equation}

Preference construction for hallucination mitigation requires pairs that contrast visually faithful responses with image-unsupported ones. Intervention-based methods create such contrasts by externally modifying responses, but this can move the data away from the policy distribution. Sampling from the policy reduces this mismatch; however, even when the policy is conditioned on the image $\vm$ in Eq.~\ref{eq:policy}, its rollouts may still be driven by language priors or modality biases rather than image-specific evidence. We therefore aim to construct preference pairs that remain close to the policy distribution while using object-level visual cues extracted from the image as additional conditioning for the policy's own rollouts.

As summarized in Figure~\ref{fig:overview}, \methodname{} achieves this goal by extracting a consistent visual cue from semantically aligned image variations, as detailed in Section~\ref{method:self-caption}. The key intuition is that visual content present in the original image should recur across synthetic images generated from diverse captions describing the same image. This consistent visual cue is then used to guide cue-conditioned self-rollouts from the policy, as described in Section~\ref{method:cue-conditioned}. Rather than constructing preferences from externally generated responses, we sample cue-conditioned self-generated candidate responses from the policy itself. Thus, the visual cue encourages visual grounding, while the policy's own rollouts keep the candidates close to its response distribution.

\subsection{Self-captioned semantic synthesis for {visual} cue extraction}
\label{method:self-caption}

To construct semantically aligned synthetic images, we use the policy's captions of images as T2I prompts. Concretely, we define the captioning prompt as $\vx_{\text{cap.}} = \text{``\textit{Describe the image in detail.}''}$. We sample $N_c$ captions and synthesize their corresponding images using a T2I diffusion model $P_{\mathbf{\psi}}$:
\begin{equation}
    \{\vq_j\}_{j=1}^{N_c} \sim \pi_{\vtheta}(\cdot \mid \vx_{\text{cap.}}, \vm),
    \qquad
    \{\vm'_j\}_{j=1}^{N_c} = \{P_{\mathbf{\psi}}(\vq_j)\}_{j=1}^{N_c}.
\end{equation}

Although the sampled captions are grounded in the same image, they emphasize different aspects of the image. Since these images share the same underlying semantic content, comparing them against the original image helps identify image-supported evidence.

\paragraph{Visual cue extraction}
From the self-captioned synthetic images $\vm'_j$, we extract a pairwise cue $\vc_j$ for each $\vm'_j$ by comparing it with the original image. Specifically, for the synthetic image indexed by $j$, the judge model extracts object-level visual evidence that is consistent between the original image $\vm$ and the corresponding synthetic image $\vm'_j$:
\begin{equation}
\vc_j = \pi_{\bm{\phi}}(\vx_{\text{comp.}}, \vm, \vm'_j),
\end{equation}
where $\bm{\phi}$ denotes a judge model and $\vx_{\text{comp.}}$ is the pairwise comparison prompt provided in Appendix~\ref{ap:prompt}. The pairwise cues are then aggregated into a consistent visual cue:
\begin{equation}
\vc = \pi_{\bm{\phi}}(\vx_{\text{agg.}}, \{\vc_j\}_{j=1}^{N_c}),
\end{equation}

where $\vx_{\text{agg.}}$ is the aggregation prompt provided in Appendix~\ref{ap:prompt}. Since T2I introduces variations in visual appearance, cues that recur consistently across multiple pairwise comparisons are more likely to reflect content grounded in the original image. The aggregation combines these pairwise consistent cues to produce the consistent visual cue $\vc$, which serves as guidance for preference construction.

\subsection{Cue-conditioned self-distillation for preference synthesis}

\label{method:cue-conditioned}
The visual cue $\vc$ guides the policy's own rollouts. We sample $N_s$ cue-conditioned self-generated candidate responses from the policy:
\begin{equation}
\vy^{(s)} \sim \pi_{\vtheta}(\cdot \mid \vx_{\text{cue-cap.}}, \vm, \vc),
\end{equation}
where $s$ indexes the sampled candidate responses and $\vx_{\text{cue-cap.}}$ denotes the cue-conditioned captioning prompt provided in Appendix~\ref{ap:prompt}. Repeated sampling under the same prompt, image, and visual cue produces diverse cue-conditioned candidate responses. Since these candidates are sampled from the policy itself, they remain close to the policy's response distribution, while the visual cue encourages them to leverage visual information from the image, resulting in varying degrees of visual grounding across candidates.

\paragraph{Preference selection by visual grounding}
From the cue-conditioned candidate responses $\{\vy^{(s)}\}_{s=1}^{N_s}$, we select a preferred response $\vy_w$ and a dispreferred response $\vy_l$ based on visual grounding. Specifically, we query the judge model with a visual grounding prompt:
\begin{equation}
(\vy_w, \vy_l) = \pi_{\bm{\phi}}\left(\vx_{\text{vis.}}, \vm, \{\vy^{(s)}\}_{s=1}^{N_s}\right),
\end{equation}
where $\vx_{\text{vis.}}$ is the visual grounding prompt provided in Appendix~\ref{ap:prompt}. The prompt asks the judge model to compare the candidate responses with respect to the original image $\vm$. The selected pair contrasts a more visually grounded response with a less grounded one, forming a policy-aligned preference pair from the policy's own cue-conditioned generations. We provide the \methodname{} algorithm in Appendix~\ref{ap:algorithm}.

\subsection{Preference optimization with \methodname{}}

Once \methodname{} constructs the preference dataset $\mathcal{D}_{\text{pref}} = (\vx_i, \vm_i, \vy_{w,i}, \vy_{l,i})$, the pairs can be used with a range of pairwise preference optimization objectives \cite{zhao2023slic,azar2024general,hong2024orpo,meng2024simpo}. We instantiate this step with DPO \cite{rafailov2023direct} to match the optimization objective used in recent hallucination mitigation methods, enabling a fair comparison while isolating the effect of our preference construction. We optimize:
\begin{align}
\mathcal{L}_{\text{DPO}}
= - \mathbb{E}_{(\vx, \vm, \vy_w, \vy_l) \sim \mathcal{D}_{\text{pref}}}
\left[
\log \sigma \left(
\beta \log \frac{\pi_\vtheta(\vy_w \mid \vx, \vm)}{\pi_{\text{ref}}(\vy_w \mid \vx, \vm)}
-
\beta \log \frac{\pi_\vtheta(\vy_l \mid \vx, \vm)}{\pi_{\text{ref}}(\vy_l \mid \vx, \vm)}
\right)
\right],
\label{eq:dpo}
\end{align}
where $\pi_{\text{ref}}$ denotes the reference policy and $\beta$ is a scaling parameter.
The consistent visual cue $\vc$ and self-captioned synthetic images $\{\vm'_j\}_{j=1}^{K}$ are used only for constructing the preference pairs, not as inputs to the policy or reference model during DPO. During optimization, the policy and reference model evaluate the constructed responses $(\vy_w, \vy_l)$ conditioned only on the original prompt-image pair $(\vx, \vm)$. This design enables cue-conditioned self-distillation by using preference optimization to distill cue-guided visual grounding into the image-conditioned policy. Consequently, \methodname{} improves visual grounding without requiring cue extraction or synthetic image generation at inference time.

\begin{table*}[t]
\centering
\setlength{\tabcolsep}{3.5pt}
\caption{
Comparison on hallucination and general benchmarks using LLaVA-1.5-7B \& LLaVA-1.5-13B \cite{liu2024improved}. \textbf{Bold} and \underline{underline} values indicate the best and second-best results, respectively.
}
\begin{adjustbox}{width=\textwidth}
\begin{tabular}{@{}c l| cc ccc c |c c c@{}}
\toprule

& 
& \multicolumn{6}{c|}{\textbf{Hallucination benchmarks}}
& \multicolumn{3}{c}{\textbf{General benchmarks}} \\

\cmidrule(lr){3-8} \cmidrule(lr){9-11}

\textbf{Model}
& \textbf{Method}
& \multicolumn{2}{c}{Object HalBench}
& \multicolumn{3}{c}{AMBER}
& HallusionBench
& MMStar 
& MMVP 
& CV-Bench  \\

\cmidrule(lr){3-4} \cmidrule(lr){5-7} \cmidrule(lr){8-8}
\cmidrule(lr){9-9} \cmidrule(lr){10-10} \cmidrule(lr){11-11}

&
& Resp.$\downarrow$ & Ment.$\downarrow$
& CHAIR$\downarrow$ & Hal.$\downarrow$ & Cog.$\downarrow$
& aAcc.$\uparrow$
& Acc.$\uparrow$ & Acc.$\uparrow$ & Acc.$\uparrow$ \\

\midrule

\multirow{14}{*}{LLaVA-1.5-7B}
& Vanilla
& 51.3 & 25.9
& 7.9 & 36.8 & 4.3
& 46.8
& 33.4 & 23.3 & 61.9 \\

\cmidrule(lr){2-11}

& VCD \cite{leng2024mitigating}
& 50.6 & 25.6 & 8.1 & 37.9 & 4.0 & 47.2 & \underline{34.5} & 25.3 & 56.0 \\

& DOLA \cite{chuang2023dola}
& 44.3 & 24.9 & 6.5 & 28.5 & 3.1 & 46.7 & 33.5 & 22.6 & 62.0 \\

& OPERA \cite{huang2024opera}
& 42.0 & 22.6 & 5.8 & 27.7 & 3.0 & 45.7 & 33.4 & 22.6 & 61.9 \\

& EFUF \cite{xing2024efuf}
& 41.6 & 22.6
& 5.8 & 28.2 & 3.0
& 47.0
& 33.6 & 24.6 & 62.2 \\

& LLaVA-RLHF \cite{sun2024aligning}
& 52.6  & 26.9  & 9.8   & 46.4  & 5.4   & 41.8  & 20.1  & 12.0    & 27.8 \\

& HALVA \cite{sarkar2024halva}
& 41.3 & 20.5
& 6.8 & 33.3 & 3.5
& \textbf{48.8}
& 32.7 & 23.3 & 61.3 \\

& HA-DPO \cite{zhao2023beyond}
& 39.0 & 19.6
& 6.6 & 30.9 & 3.2
& \underline{48.1}
& 32.7 & 22.6 & 61.4 \\

& POVID \cite{zhou2024aligning}
& 31.6 & 16.3
& 5.4 & 28.7 & 3.0
& 46.5
& 33.6 & 25.3 & 62.6 \\

& TPR \cite{he2024systematic}
& 11.0 & 5.8
& 3.6 & 20.4 & 1.7
& 39.8
& 33.2 & \underline{29.3} & \underline{63.6} \\

& RLAIF-V \cite{yu2024rlaif}
& 8.6 & 4.1
& 2.9 & 15.8 & \underline{0.9}
& 35.1
& {34.0} & 24.0 & 42.4 \\

& OPA-DPO \cite{yang2025opadpo}
& 8.6 & 5.1
& \underline{2.7} & 13.7 & 1.1
& 47.9
& 32.8 & 28.0 & 59.5 \\

& SENTINEL \cite{peng2025mitigating}
& \underline{5.3} & \underline{3.0}
& 2.8 & \underline{13.6} & 1.2
& 47.6
& 33.6 & 24.6 & 61.9 \\

\cmidrule(lr){2-11}

\rowcolor{cyan!15}
& ViPSy (\textbf{Ours})
& \textbf{4.0}
& \textbf{2.6}
& \textbf{1.8}
& \textbf{9.6}
& \textbf{0.6}
& \underline{48.1}
& \textbf{35.4}
& \textbf{30.6}
& \textbf{64.4} \\

\midrule

\multirow{7}{*}{LLaVA-1.5-13B}
& Vanilla
& 44.6 & 22.1 & 5.9 & 28.1 & 4.1 & 46.5 & 33.2 & 32.0 & 61.7 \\

\cmidrule(lr){2-11}

& HALVA \cite{sarkar2024halva}
& 47.3 & 12.8& 5.8 & 28.8 & 4.1 & \underline{47.6} & 33.4 & 32.6 & \underline{61.4} \\

& LLaVA-RLHF \cite{sun2024aligning}
& 36.3  & 11.8 & 6.8   & 34.1  & 5.2   & 45.9  & 20.5  & 21.3 & 37.6\\

& HSA-DPO \cite{xiao2025detecting}
& {5.0} & {3.0} & 2.2 & 12.2 & 1.6 & 46.1 & 34.5 & \underline{33.3} & 60.1 \\

& OPA-DPO \cite{yang2025opadpo}
& 6.0 & 3.7 & \underline{2.1 }& \underline{9.4} & 1.4 & 46.4 & \underline{34.6} & 31.3 & 61.3 \\

& SENTINEL \cite{peng2025mitigating}
& \underline{3.3} & \underline{1.9} & 2.4 & 11.9 & \underline{1.2} & 47.4 & 32.4 & \textbf{35.3} & 60.5 \\

\cmidrule(lr){2-11}

\rowcolor{cyan!15}
& ViPSy (\textbf{Ours})
&\textbf{2.6}   & \textbf{1.3}   &\textbf{1.9}   & \textbf{9.2}   & \textbf{1.1}   & \textbf{48.0}    & \textbf{35.8}  & \textbf{35.3}  & \textbf{62.6} \\

\bottomrule
\end{tabular}
\end{adjustbox}
\label{tab:main_table_llava}
\end{table*}

\section{Experiments}
\label{sec:experiments}

Our experiments evaluate whether \methodname{} addresses the two limitations identified above: (a) policy-distribution mismatch in intervention-based methods and (b) limited use of visual information in sampling-based preference construction. To validate it, we assess whether the constructed preference pairs reduce hallucination while enhancing general vision-language capability, and further examine whether self-derived visual cues strengthen the vision encoder through visual representation and segmentation evaluations rather than merely suppressing language-prior-driven hallucination.

\subsection{Setups}
\label{experiments}

\paragraph{Models}
Following prior work, we use LLaVA-1.5-7B and LLaVA-1.5-13B \cite{liu2024improved} as the policy models for direct comparison with existing hallucination mitigation methods. We use Qwen2.5-VL-72B \cite{qwen2.5-VL} as the default judge model and SD-3.5-L \cite{esser2024scaling} as the default T2I diffusion model for semantic synthesis. To test whether the proposed preference construction is policy-agnostic, we further evaluate diverse VLMs, including Qwen2.5-VL-7B \cite{qwen2.5-VL}, InternVL3.5-8B \cite{wang2025internvl3}, mPLUG-Owl3-7B \cite{ye2024mplug}, and Kimi-VL-A3B \cite{kimiteam2025kimivltechnicalreport}. We also consider Qwen3-VL-30B \cite{bai2025qwen3} and GPT-5 mini \cite{singh2025openai} as alternative judges, and SD-3.5-M \cite{esser2024scaling}, Flux-1.0-dev \cite{flux2024}, and PixelDiT \cite{yu2025pixeldit} as alternative T2I models in ablations. Additional details on policy, judge, and diffusion models are provided in Appendix~\ref{ap:model}.

\paragraph{Baselines}
We compare \methodname{} with hallucination-mitigation baselines, including decoding methods \cite{leng2024mitigating, chuang2023dola, huang2024opera}, which work only at inference time, and unlearning-based methods \cite{xing2024efuf}, which use curated data. Intervention-based methods \cite{zhao2023beyond, zhou2024aligning, sarkar2024halva, he2024systematic, xiao2025detecting} can deviate from the policy distribution, while sampling-based methods \cite{sun2024aligning, yu2024rlaif, yang2025opadpo, peng2025mitigating} often underuse visual information during preference construction. Detailed descriptions are provided in Appendix~\ref{ap:baseline}.

\paragraph{Datasets and evaluation}
We construct preference data from Visual Genome images \cite{krishna2017visual}, following \citet{peng2025mitigating}, with implementation and training details in Appendix~\ref{ap:experimental_details}. We evaluate hallucination on Object HalBench \cite{rohrbach2018object}, AMBER \cite{wang2023amber}, and HallusionBench \cite{guan2024hallusionbench}, and general vision-language capability on MMStar \cite{chen2024we}, MMVP \cite{tong2024eyes}, and CV-Bench \cite{tong2024cambrian}. For vision encoder evaluation, we use ImageNet 1k \cite{imagenet15russakovsky}, ImageNet-O \cite{hendrycks2021nae}, and Rendered-SST2 \cite{socher2013recursive, radford2021learning} for linear probing, and MS COCO \cite{lin2014microsoft} for segmentation. Dataset and metric details are provided in Appendix~\ref{ap:data}.

\begin{figure*}[t]
  \centering
  \makebox[\textwidth][c]{
    \begin{minipage}[t]{0.285\textwidth}
      \vspace{0pt}
      \centering
    \captionof{table}{Average hallucination and general benchmark scores across policy models.}

      \resizebox{\linewidth}{!}{
        \begin{tabular}{@{}l| c c@{}}
\toprule

\textbf{Method} 
& \textbf{Hal. Avg. $\downarrow$}
& \textbf{Gen. Avg. $\uparrow$} \\

\midrule

\rowcolor{gray!15}
\multicolumn{3}{@{}c@{}}{\textbf{Qwen2.5-VL-7B}} \\
\midrule
Vanilla     
& 9.48  & 64.90 \\
\rowcolor{cyan!15}
ViPSy     
& \textbf{8.16}  & \textbf{66.10} \\

\midrule

\rowcolor{gray!15}
\multicolumn{3}{@{}c@{}}{\textbf{mPLUG-Owl3-7B}} \\
\midrule
Vanilla     
& 6.66  & 48.87 \\
\rowcolor{cyan!15}
ViPSy     
& \textbf{5.84}  & \textbf{50.10} \\

\midrule

\rowcolor{gray!15}
\multicolumn{3}{@{}c@{}}{\textbf{InternVL3.5-8B}} \\
\midrule
Vanilla     
& 8.66  & 69.07 \\
\rowcolor{cyan!15}
ViPSy     
& \textbf{6.78}  & \textbf{69.77} \\

\midrule

\rowcolor{gray!15}
\multicolumn{3}{@{}c@{}}{\textbf{Kimi-VL-A3B}} \\
\midrule
Vanilla     
& 11.52 & 57.97 \\
\rowcolor{cyan!15}
ViPSy  
& \textbf{8.70}   & \textbf{59.53} \\

\bottomrule
\end{tabular}

      }
      \label{tab:policy_ablation}
    \end{minipage}
    \hspace{0.02\textwidth}
    \begin{minipage}[t]{0.65\textwidth}
      \vspace{0pt}
      \centering
    \captionof{table}{Ablation study on model configurations based on average scores across hallucination~\cite{rohrbach2018object, wang2023amber} and general benchmarks~\cite{chen2024we, tong2024eyes, tong2024cambrian}, respectively. ``Self'' denotes sampling from LLaVA-1.5-7B.}

      \resizebox{\linewidth}{!}{
        \begin{tabular}{@{}ccc|cc@{}}
\toprule

\multicolumn{3}{@{}c|}{\textbf{\methodname{} Configuration}}
& \multicolumn{2}{c}{\textbf{Benchmarks}} \\

\cmidrule(lr){1-3}
\cmidrule(lr){4-5}

\textbf{I$\rightarrow$T}
& \textbf{T$\rightarrow$I}
& \textbf{Rollout}
& \textbf{Hal. Avg. $\downarrow$}
& \textbf{Gen. Avg. $\uparrow$} \\

\midrule

\rowcolor{red!10}
\multicolumn{3}{@{}l|}{\textit{Vanilla (LLaVA-1.5-7B)}}
& 25.24
& 39.53 \\

\midrule

\rowcolor{cyan!15}
Self
& SD-3.5-L 
& Self
& \textbf{3.72}
& \textbf{43.47} \\

\midrule
GPT-5 mini
& \textcolor{lightgray}{SD-3.5-L}
& \textcolor{lightgray}{Self}
& 5.12
& 42.57 \\

Qwen3-VL-30B
& \textcolor{lightgray}{SD-3.5-L}
& \textcolor{lightgray}{Self}
& 4.70
& 42.13 \\

Qwen2.5-VL-72B
& \textcolor{lightgray}{SD-3.5-L}
& \textcolor{lightgray}{Self}
& 4.26
& 42.40 \\

\midrule

\textcolor{lightgray}{Self}
& PixelDiT 
& \textcolor{lightgray}{Self}
& 4.50
& 42.70 \\

\textcolor{lightgray}{Self}
& Flux-1.0-dev 
& \textcolor{lightgray}{Self}
& 4.36
& 42.50 \\

\textcolor{lightgray}{Self}
& SD-3.5-M 
& \textcolor{lightgray}{Self}
& 3.86
& 42.87 \\

\midrule

\textcolor{lightgray}{Self}
& \textcolor{lightgray}{SD-3.5-L}
& GPT-5 mini
& 15.64
& 41.07 \\

\textcolor{lightgray}{Self}
& \textcolor{lightgray}{SD-3.5-L}
& Qwen3-VL-30B
& 16.00
& 42.07 \\

\textcolor{lightgray}{Self}
& \textcolor{lightgray}{SD-3.5-L}
& Qwen2.5-VL-72B
& 11.22
& 41.80 \\

\bottomrule
\end{tabular}
      }
      \vspace{0pt}
      \label{tab:ablation_roll_t2i}
    \end{minipage}
  }
  \vspace{-0.05in}
\end{figure*}

\begin{table*}[t]
  \centering
\caption{
Vision encoder evaluation with 2-, 4-, and 8-shot linear probing accuracy on ImageNet-1k (IN-1k), ImageNet-O (IN-O), and Rendered-SST2 (R-SST2), and segmentation recall on MS COCO.
}

  \begin{adjustbox}{max width=\textwidth}
    \begin{tabular}{lcccccccccc}
\toprule

& \multicolumn{9}{c}{Linear Probing}
& \multicolumn{1}{c}{Seg.} \\

\cmidrule(lr){2-10} \cmidrule(lr){11-11}

\textbf{Method}
& \multicolumn{3}{c}{2-shot}
& \multicolumn{3}{c}{4-shot}
& \multicolumn{3}{c}{8-shot}
&  \\

\cmidrule(lr){2-4}
\cmidrule(lr){5-7}
\cmidrule(lr){8-10}
\cmidrule(lr){11-11}

& IN-1k
& IN-O
& R-SST2
& IN-1k
& IN-O
& R-SST2
& IN-1k
& IN-O
& R-SST2
& MS COCO \\

& Acc.$\uparrow$
& Acc.$\uparrow$
& Acc.$\uparrow$
& Acc.$\uparrow$
& Acc.$\uparrow$
& Acc.$\uparrow$
& Acc.$\uparrow$
& Acc.$\uparrow$
& Acc.$\uparrow$
& Recall $\uparrow$ \\

\midrule

Vanilla

& 35.4  & 18.0    & 48.9  & 48.6  & 29.5  & 50.4  & 59.3  & 38.0    & 50.1  & 34.7 \\

\rowcolor{cyan!15}
ViPSy (\textbf{Ours})
& \textbf{36.1}  & \textbf{18.3}  & \textbf{49.5}  & \textbf{49.0}    & \textbf{30.0}    &\textbf{51.0}    & \textbf{60.0}    & \textbf{38.5}  & \textbf{50.9}  & \textbf{35.9} \\
\bottomrule
\end{tabular}

  \end{adjustbox}
  \label{tab:vision_encoder}
  \vspace{-0.15in}
\end{table*}

\subsection{Main results}

\paragraph{Mitigating hallucination}
Table~\ref{tab:main_table_llava} shows that \methodname{} achieves state-of-the-art hallucination mitigation on LLaVA-1.5-7B and remains effective on LLaVA-1.5-13B. On LLaVA-1.5-7B, against TPR~\cite{he2024systematic}, the strongest intervention-based method, our method reduces Object HalBench Resp./Ment. by 63.6\%/55.2\% and AMBER CHAIR/Hal./Cog. by 50.0\%/52.9\%/64.7\%. Against the strongest sampling-based, SENTINEL~\cite{peng2025mitigating}, it improves Object HalBench Resp./Ment. by 24.5\%/13.3\% and AMBER CHAIR/Hal./Cog. by 35.7\%/29.4\%/50.0\%. These gains support policy-aligned, visually grounded preference construction. Although it ranks second on HallusionBench, it achieves the best results on the remaining benchmarks, showing broader gains in faithful visual understanding. On the larger LLaVA-1.5-13B model, \methodname{} outperforms all baselines across all hallucination benchmarks. In particular, compared with the previous state-of-the-art, it achieves notable improvements of 21.2\%/31.6\% on Object HalBench Resp./Ment. and 20.8\%/22.7\%/8.3\% on AMBER CHAIR/Hal./Cog. These results indicate that \methodname{} remains effective as the model scales. Qualitative examples in Appendix~\ref{ap:mitigate_hal_qual} show that preference-aligned descriptions using preference pairs constructed by our method avoid unsupported details and produce image-grounded descriptions.

\paragraph{General performance}
On general benchmarks with LLaVA-1.5-7B, our preference construction improves over vanilla on all three benchmarks, with gains of 6.0\% on MMStar, 31.3\% on MMVP, and 4.0\% on CV-Bench, achieving the best overall results among compared methods. Unlike sampling-based methods that degrade or merely maintain CV-Bench, our method improves it, suggesting that visual cues enhance general vision-language capability. Together with the hallucination results, these gains support policy-aligned, visually grounded preference pairs. The results on LLaVA-1.5-13B further demonstrate scalability, as our method outperforms all compared methods across all benchmarks and, in particular, improves over vanilla by 7.8\% on MMStar and 10.3\% on MMVP.

\paragraph{Generalization across policy models}
The effectiveness of \methodname{} is not specific to LLaVA-1.5-7B. Table~\ref{tab:policy_ablation} shows that our method consistently generalizes across diverse policy models, improving both hallucination mitigation and general vision-language capabilities. Compared with the vanilla policy, it reduces the hallucination average, computed over two hallucination benchmarks~\cite{rohrbach2018object, wang2023amber}, by 13.9\% on Qwen2.5-VL-7B, 12.3\% on mPLUG-Owl3-7B, 21.7\% on InternVL3.5-8B, and 24.5\% on Kimi-VL-A3B. At the same time, it improves the general average across three general benchmarks~\cite{chen2024we, tong2024eyes, tong2024cambrian}. These results demonstrate that policy-aligned, visually grounded preference construction provides consistent benefits beyond a single policy model. Full results are provided in Appendix~\ref{ap:policy_abl}.

\begin{figure*}[t]
  \centering
  \includegraphics[width=\linewidth]{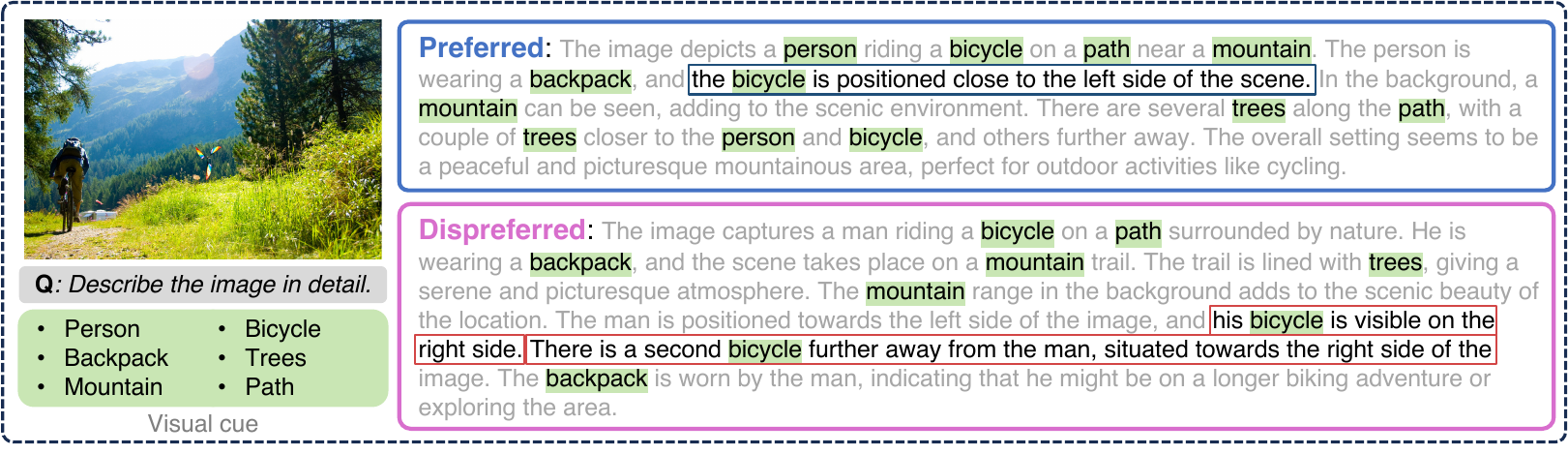}

    \caption{Qualitative examples of cue-conditioned preference pairs. Given the same image, prompt, and visual cue, preferred responses remain image-grounded, while dispreferred ones hallucinate. Additional examples are shown in Appendix~\ref{ap:preference_data_examples}, \ref{ap:mitigate_hal_qual}, and \ref{ap:self-cap-synthesis_qual}: preference pairs, aligned-policy image descriptions, and self-captioned semantic synthesis, respectively.}
    \vspace{-0.15in}
    \label{fig:cue_rollout}
\end{figure*}

\subsection{Analysis}

\paragraph{Effect on vision encoder}
To examine whether our approach improves the vision encoder beyond response-level hallucination mitigation, we use few-shot linear probing and segmentation. We update the vision encoder during preference alignment and freeze the resulting encoder for linear probing and segmentation. Following \citet{liu2025visual}, we freeze the resulting vision encoder and train dataset-specific linear classifiers on frozen image features under 2-, 4-, and 8-shot settings. For segmentation, we train a two-layer transformer output head to predict patch-level semantic labels, following \citet{covert2024locality}. As shown in Table~\ref{tab:vision_encoder}, our method improves ImageNet-1k, ImageNet-O, and Rendered-SST2 accuracy as well as segmentation recall over vanilla. These gains suggest that \methodname{} improves both visual representation quality and the ability to separate visual regions, rather than merely suppressing language-prior-driven hallucinations in responses.

\paragraph{Qualitative analysis of cue-conditioned rollouts}
We show how \methodname{} obtains preferred and dispreferred responses, and we present examples of preference pairs selected from multiple cue-conditioned rollouts under the same image, prompt, and visual cue. Preferred responses use the visual cue to ground the image description, whereas dispreferred responses use the cue but add details unsupported by the image. As highlighted in Figure~\ref{fig:cue_rollout}, the same visual cue can yield candidates with different grounding quality, e.g., in the bicycle scene, the preferred response correctly locates the bicycle close to the left side of the scene, while the dispreferred response states that the bicycle is on the right side and further hallucinates a second bicycle, although only one bicycle is visible. This demonstrates that preference selection is needed to separate faithful cue usage from unsupported extrapolation. More examples are provided in Appendix~\ref{ap:preference_data_examples}. 

\paragraph{Judge agreement and robustness}
Since preference selection is performed by a judge over multiple cue-conditioned rollouts, one concern is whether the selected pairs depend on the judge model. We therefore evaluate the agreement between the default judge, Qwen2.5-VL-72B, and alternative judges. We also measure hallucination and general capabilities when each judge is used to construct preference pairs. As shown in Figure~\ref{fig:reference_agreement}, Qwen3-VL-30B and GPT-5 mini agree with the default judge on 98.4\% and 95.8\% of preference selections, respectively, indicating that the selected preferences are highly consistent across judges. Furthermore, Table~\ref{tab:ap_judge_ablation} shows that using different judges to construct preference pairs leads to similar results on both hallucination and general benchmarks. This suggests that the preference construction process is not dominated by judge-specific bias and that \methodname{} reliably constructs visually grounded preference pairs. Detailed results are provided in Appendix~\ref{ap:selection_judge_ablation}.

\subsection{Ablation study}

\paragraph{Visual cue construction}
We analyze visual-cue construction by varying the captioning and T2I models. The input image is captioned, and the caption is used to synthesize an image for cue extraction. As shown in Table~\ref{tab:ablation_roll_t2i}, self-captioning achieves the best overall result, while alternative T2I models remain competitive with moderate performance differences. This suggests that using policy-generated captions is beneficial for cue construction, and that \methodname{} is not overly sensitive to the T2I model. Full results are in Appendix~\ref{ap:model_abl}, and Qualitative examples in Appendix~\ref{ap:self-cap-synthesis_qual} show that self-captioned synthetic images preserve the main visual context while varying across T2I models.

\paragraph{Rollout model}
We compare models for cue-conditioned rollout generation while keeping the visual-cue construction fixed. Self-rollouts achieve the best hallucination score and general performance, whereas rollouts from external VLMs lead to degraded results. This shows that candidate responses should remain close to the policy distribution while being guided by visual cues. Full results are provided in Appendix~\ref{ap:model_abl}.

\begin{figure*}[t]

  \centering

  \begin{minipage}[t]{0.34\textwidth}
    \vspace{5pt}
    \centering
    \begin{minipage}[t][4.2cm][c]{\linewidth}
      \centering
      \includegraphics[width=\linewidth ]{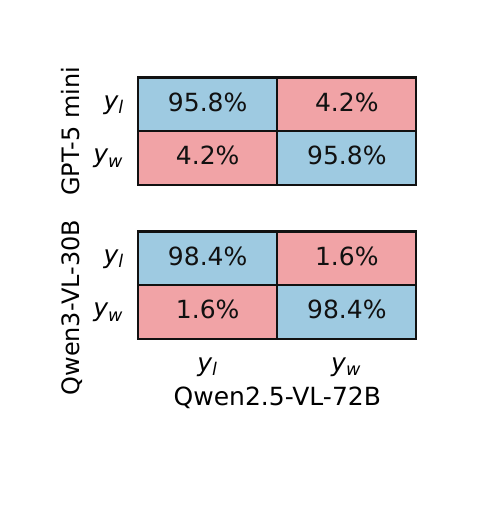}
    \end{minipage}
    \vspace{-16.8pt}
    \captionof{figure}{Agreement of preference selections for alternative judges compared upon preferences based on Qwen2.5-VL-72B.}
    \label{fig:reference_agreement}
  \end{minipage}
  \hfill
  \begin{minipage}[t]{0.3\textwidth}
    \vspace{-0pt}
    \centering
    \begin{minipage}[t][4.2cm][c]{\linewidth}
      \centering
      \includegraphics[width=\linewidth ]{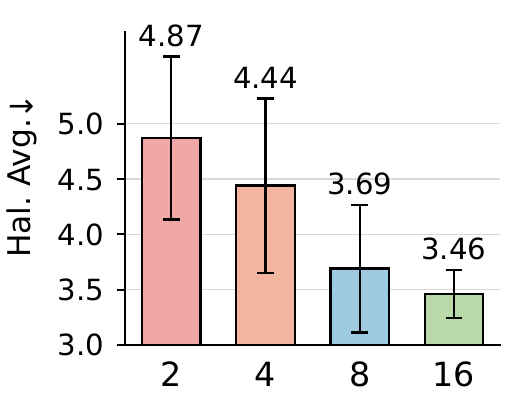}
    \end{minipage}
    \vspace{-11pt}
    \captionof{figure}{Effect of varying the number of cue-conditioned rollouts on average hallucination errors across five runs.}
    \label{fig:num_rollouts}
  \end{minipage}
  \hfill
  \begin{minipage}[t]{0.3\textwidth}
    \vspace{0pt}
    \centering
    \begin{minipage}[t][4.2cm][c]{\linewidth}
      \centering
      \resizebox{\linewidth}{!}{
        \begin{tabular}{@{}l| c c@{}}
\toprule

\textbf{Strategy}
& \textbf{Hal. Avg. $\downarrow$}
& \textbf{Gen. Avg. $\uparrow$} \\

\midrule
\rowcolor{red!10}
Vanilla     
& 25.24 & 39.53 \\

\midrule
w/o cue
&7.14 & 42.50 \\
\midrule

Translation
& 5.10   & 43.13 \\

Resize
&4.82  & 42.47 \\

Mask
&4.82  & 42.53 \\

Crop
& 4.76  & 42.43 \\

Video
& 4.46  & 41.70 \\

\midrule
\rowcolor{cyan!15}
Diffusion
& \textbf{3.72}  & \textbf{43.47} \\

\bottomrule
\end{tabular}

      }
    \end{minipage}

    \vspace{-9.7pt}

    \captionof{table}{Comparison of visual variations for cue extraction in average hallucination and general benchmark scores.}
    \label{tab:various_synthesis}
  \end{minipage}
\vspace{-0.15in}
\end{figure*}

\paragraph{Number of rollouts}
To assess how the number of rollouts affects preference construction, we vary the number of cue-conditioned rollouts. As shown in Figure~\ref{fig:num_rollouts}, more rollouts improve reliability by enlarging the candidate pool and increasing grounding-quality diversity. Increasing the number of rollouts reduces variance in hallucination score, as the larger candidate pool makes the preferred and dispreferred responses more clearly distinguishable. Since more rollouts can improve preference construction but increase computational cost, we use 8 cue-conditioned rollouts as a trade-off between reliability and efficiency. Additional results are included in Appendix~\ref{ap:num_rollout_ablation}.

\paragraph{Cue conditioning and synthesis strategy}
Table~\ref{tab:various_synthesis} studies cue conditioning and visual variation strategies for cue extraction. Even without cues, policy-sampled candidates improve over the vanilla model, but cue-conditioned construction performs better, confirming the importance of visual cues for visually faithful rollouts. Among cue extraction strategies, simple transformations such as translation, resize, mask, and crop bring improvements but provide limited semantic diversity. Video-based variation using an image-to-video model~\cite{wan2.1} can introduce uncontrolled changes. Diffusion-based synthesis produces richer semantic variations and achieves the best overall result, with the lowest hallucination score and the highest general capability. Detailed results are reported in Appendix~\ref{ap:synthesis_strategy}.

\section{Conclusion}
\label{sec:conclusion}

In this paper, we proposed \methodname{}, a vision-driven preference synthesis that constructs preference pairs leveraging visual information while remaining aligned with the policy distribution. Unlike intervention-based methods that externally modify responses and sampling-based methods that overlook visual information, our approach first extracts visual cues by comparing the original image with semantically synthesized image variations. These cues then guide the policy model's own rollouts, allowing candidate responses to remain close to the model's generation distribution while encouraging them to leverage visual information from the input image. Experiments show that our approach achieves state-of-the-art hallucination mitigation while also improving general visual grounding benchmarks, with additional gains observed in vision encoder evaluations.

\bibliographystyle{unsrtnat}
{\small

\bibliography{main}
}

\clearpage

\appendix

\section{Limitations and broader impact}
\label{ap:limitation_broad_impact}

Although \methodname{} achieves strong hallucination mitigation, it relies on external models for preference construction, including a text-to-image diffusion model and a judge VLM. Consequently, the resulting preference pairs may inherit biases or failure modes from these models when synthesized variants or judge decisions do not faithfully reflect image-grounded evidence. Improving or domain-adapting these external models could further enhance preference quality. This study also focuses mainly on object-level visual grounding with cues from Visual Genome images. While effective for reducing object hallucination, broader scenarios such as complex reasoning, spatial understanding, and specialized domains remain underexplored. \methodname{} may improve VLM reliability in applications such as education, accessibility, robotics, industrial inspection, and assistive scientific or medical workflows. 

\section{Training details}
\label{ap:experimental_details}

We build our experiments on top of LLaVA-1.5 \cite{liu2024improved}, using Vicuna-1.5 \cite{vicuna2023} as the language model backbone and CLIP ViT-L/336px \cite{radford2021learning} as the vision encoder. The visual features are mapped to the language space through the default two-layer GELU MLP projector. The training set is constructed from Visual Genome images \cite{krishna2017visual}. Our final preference dataset contains 7K preference pairs, which is smaller than the training data used by \citet{peng2025mitigating}, the previous state-of-the-art method for hallucination mitigation. This allows us to evaluate whether visually grounded preference construction can be effective even with a relatively compact preference dataset. For optimization, we use LoRA-based parameter-efficient tuning. The tuning targets include the LLM, the vision encoder, and the projector. We insert trainable LoRA modules into these components rather than updating all model parameters. The LoRA modules for the LLM and projector are trained with a learning rate of $2\times10^{-5}$, while the vision encoder uses $2\times10^{-6}$. We use LoRA rank $r=128$, scaling factor $\alpha=256$, and set the preference loss coefficient to $\beta=0.1$. Models are trained with AdamW, cosine learning-rate decay, no weight decay, and a maximum context length of 2048. Training is performed for one epoch with a global batch size of 64. We employ ZeRO stage-2 optimization to reduce memory usage. Both $N_c$ and $N_s$ are set to 8, representing the number of captions and cue-conditioned self-generated candidates, respectively.

\paragraph{Compute infrastructure}

All experiments were conducted on compute cluster instances with two NVIDIA B200 GPUs, each with 183GB of VRAM, two Intel Xeon 6960P CPUs with 144 physical cores and 288 threads in total.

\section{Models}
\label{ap:model}

\paragraph{Policy Models}

\begin{itemize}
    \item \textbf{LLaVA 1.5}~\cite{liu2024improved} is used as the main policy model in our experiments. We evaluate both the 7B and 13B variants to study the effect of policy model scale. LLaVA 1.5 combines a CLIP-based visual encoder with a Vicuna language model through a multimodal projector, enabling it to generate language responses conditioned on both images and textual instructions. In our framework, LLaVA 1.5 generates self-captions, produces cue-conditioned candidate responses, and is optimized with the preference pairs constructed by \methodname{}. Its relatively simple connector-based design also makes it a useful baseline for isolating how preference optimization, rather than architectural complexity, affects visual grounding behavior.
    
    \item \textbf{InternVL3.5-8B}~\cite{wang2025internvl3} is used as an additional policy model to evaluate whether \methodname{} generalizes beyond the LLaVA family. InternVL3.5 is a multimodal instruction-following model designed for visual understanding and reasoning. Including this model allows us to test whether visually grounded preference construction remains effective for a different VLM architecture. 
    
    \item \textbf{Kimi-VL-A3B}~\cite{kimiteam2025kimivltechnicalreport} is used as a policy model for evaluating generalization across model families. It is a vision-language model developed for multimodal instruction following and image-grounded reasoning. This model is included to evaluate whether \methodname{} improves visual grounding when applied to a policy model with a distinct multimodal architecture.
    
    \item \textbf{mPLUG-Owl3-7B}~\cite{ye2024mplug} is used as another policy model in our policy generalization experiments. The mPLUG-Owl family follows a modular multimodal design that connects visual perception with language generation. We include this model to examine whether \methodname{} can improve visual grounding for a policy model with a different multimodal architecture. 
    
    \item \textbf{Qwen2.5-VL-7B}~\cite{qwen2.5-VL} is used as a policy model in addition to its larger judge counterpart. It is an instruction-tuned vision-language model with strong visual understanding and multimodal reasoning capabilities. Using the 7B variant as a policy model helps evaluate whether \methodname{} is effective for a compact Qwen-family VLM, rather than only for LLaVA-based policies. 
    
\end{itemize}

\paragraph{Judge Models}

\begin{itemize}
    \item \textbf{Qwen2.5-VL-72B} \cite{qwen2.5-VL} is used as the default judge model in our experiments. We use this model to extract consistent visual cues between original and synthetic images and to select preferred and dispreferred responses from cue-conditioned candidates. Due to its large scale and instruction-tuned multimodal capability, it provides a strong visual understanding for evaluating whether a response is grounded in the original image. Its fine-grained visual localization capability is useful for judging whether candidate responses preserve object- and region-level evidence rather than only global image semantics.

    \item \textbf{Qwen3-VL-30B} \cite{bai2025qwen3} is used as a judge model. It serves as an alternative multimodal evaluator for cue extraction and preference selection, allowing us to examine whether \methodname{} depends on a single judge architecture. By including Qwen3-VL-30B, we assess the robustness of the constructed preferences across different VLM judges. In our experiments, Qwen3-VL-30B is used only for judging and not as the policy model being optimized.
    
    \item \textbf{GPT-5 mini} \cite{singh2025openai} is used as an additional judge model for cue extraction and preference selection. We include it as a complementary evaluator to the Qwen-family VLM judges, allowing us to examine whether preference construction remains robust when the judge comes from a different model family. In our experiments, GPT-5 mini is used only for judging and not as the policy model being optimized. As a lightweight text-and-vision model, it provides a cost-efficient contrast to larger VLM judges while retaining multimodal reasoning ability for preference selection.
\end{itemize}

\paragraph{Text-to-Image Models}

\begin{itemize}
    \item \textbf{SD-3.5} \cite{esser2024scaling} is included as a representative latent-space diffusion model. It generates images in a compressed latent representation rather than directly in pixel space, which enables efficient image synthesis while maintaining high visual quality. In our experiments, SD-3.5-L is used to synthesize semantically aligned image variants for visual cue extraction. Its multimodal transformer backbone enables strong interaction between text and image representations, making it a reliable latent-generation baseline for cue-consistency analysis.

    \item \textbf{Flux-1.0-dev} \cite{flux2024} is included as a representative flow matching model. Unlike standard denoising diffusion formulations, flow matching learns a continuous transformation from noise to data. We use Flux-1.0-dev to test whether \methodname{} is robust to a different text-to-image generation paradigm during semantic synthesis. As a high-capacity rectified-flow transformer trained with guidance distillation, it provides a distinct synthesis backbone with strong prompt-following behavior.

    \item \textbf{PixelDiT} \cite{yu2025pixeldit} is included as a representative pixel-space diffusion model. In contrast to latent diffusion models, PixelDiT performs generation in pixel space. We include PixelDiT to examine whether visual cue extraction remains effective with pixel-space diffusion. 
\end{itemize}

\section{Baselines}
\label{ap:baseline}

We compare \methodname{} with hallucination mitigation methods for VLMs. All baselines use LLaVA-1.5 to ensure a fair comparison. We organize the baselines into three groups: sampling-based methods, intervention-based methods, and other methods. Sampling-based methods construct preference data from VLM-generated responses, which keeps the data close to the policy distribution, but they can still underuse visual information. Intervention-based methods construct preference data through external editing or correction, creating explicit contrasts between hallucinated and hallucination-free responses while moving the data away from the policy distribution. Other methods include one unlearning-based method and three training-free decoding methods.

\paragraph{Sampling-based methods}

\begin{itemize}

\item \textbf{LLaVA-RLHF}~\cite{sun2024aligning} learns a reward model from human feedback and uses reinforcement learning to train the VLM to prefer image-faithful responses. To better penalize hallucinations, the reward model is also given extra factual evidence, such as image captions or ground-truth answers.

\item \textbf{RLAIF-V}~\cite{yu2024rlaif} uses sampling-based AI feedback to reduce hallucinations in VLMs. It generates multiple candidate responses via controlled sampling under the same input, allowing preference pairs to focus on trustworthiness differences. It then scores responses at the claim level and iteratively refines the VLM using the resulting preference feedback.
    
\item \textbf{SENTINEL}~\cite{peng2025mitigating} targets object hallucination via sentence-level early intervention. It constructs in-domain preference pairs by using open-vocabulary object detectors to validate objects in VLM-generated sentences, and trains the VLM to prefer context-coherent non-hallucinated sentences over hallucinated ones, reducing hallucination propagation.

\item \textbf{OPA-DPO}~\cite{yang2025opadpo} performs on-policy alignment with corrected responses. It uses GPT-4V to minimally revise hallucinated VLM outputs and applies LoRA-SFT to align both ground-truth and revised responses with the VLM's policy distribution.

\end{itemize}

\paragraph{Intervention-based methods}

\begin{itemize}

\item \textbf{HALVA}~\cite{sarkar2024halva} addresses object hallucination in VLMs through Data-augmented Phrase-level Alignment (DPA). DPA constructs correct and hallucinated response pairs by selectively replacing grounded visual concepts in correct responses with hallucinated phrases. It penalizes hallucinated phrases relative to their correct counterparts, while using token-wise KL regularization to preserve general vision-language capabilities.

\item \textbf{HA-DPO}~\cite{zhao2023beyond} builds preference data from Visual Genome images by prompting VLMs to generate detailed descriptions and using GPT-4 with VG annotations to identify unsupported visual claims and revise them. The revised faithful responses are then paired with their original erroneous counterparts, producing positive and negative samples for preference-based training.

\item \textbf{POVID}~\cite{zhou2024aligning} frames VLM hallucination as modality misalignment and constructs preference pairs by keeping grounded responses as preferred while generating dispreferences through text- and image-side interventions. Text interventions inject plausible hallucinations, whereas image distortions trigger inherent hallucination patterns, guiding preference optimization toward visually grounded generation.

\item \textbf{HSA-DPO}~\cite{xiao2025detecting} trains a sentence-level hallucination detector from fine-grained AI feedback and uses it as a detect-then-rewrite intervention to build preference pairs. 

\item \textbf{TPR}~\cite{he2024systematic} intervenes in preference data at the topic level. Through selective replacement with VLM-resampled topic alternatives, it controls reward gaps and applies curriculum hard negative mining for data-efficient alignment.

\end{itemize}

\paragraph{Other methods}

\begin{itemize}

\item \textbf{EFUF}~\cite{xing2024efuf} is an unlearning-based method for reducing object hallucinations in VLMs. It uses CLIP-based image-relevance scores to automatically curate grounded and hallucinated object mentions without paired annotations, then unlearns hallucinated subsentences while preserving correct object generation and sentence fluency.

\item \textbf{VCD}~\cite{leng2024mitigating} is a training-free decoding method based on visual contrastive decoding. It contrasts the output distributions conditioned on the original image and a noise-distorted visual input, thereby calibrating predictions that are overly driven by language priors and statistical biases rather than visual evidence.

\item \textbf{DoLa}~\cite{chuang2023dola} is a training-free decoding method that improves factuality by contrasting predictions from earlier and later transformer layers during decoding. This helps the model rely on factual knowledge it has already learned, without using external retrieval or additional fine-tuning.

\item \textbf{OPERA}~\cite{huang2024opera} is a training-free decoding method that adjusts beam search based on self-attention patterns. It penalizes candidates that rely too much on summary tokens and, when this reliance leads to hallucination, revises the generation by returning to an earlier token and choosing a different next token.

\end{itemize}

\section{Data}
\label{ap:data}

\subsection{Training data}

We construct the VLM preference training data using images from Visual Genome (VG) \cite{krishna2017visual}. VG is a large-scale vision-language dataset that provides dense annotations for real-world images, including region descriptions, objects, attributes, relationships, and question-answer pairs. We use VG images as source images to generate visual variations and extract consistent visual cues for preference construction. Since our method relies on diverse visual scenes for this process, VG serves as a suitable data source for applying the proposed method.

\subsection{Evaluation data}

We evaluate the VLM on two groups of benchmarks, namely hallucination benchmarks and general visual-language capabilities benchmarks. In addition, to examine whether the proposed training improves the underlying vision encoder, we conduct vision encoder evaluations using linear probing and segmentation datasets. For hallucination evaluation, we use Object HalBench, AMBER, and HallusionBench. For general multimodal evaluation, we use MMStar, MMVP, and CV-Bench. For vision encoder evaluation, we use ImageNet, ImageNet-O, Rendered-SST2, and MSCOCO.

\subsubsection{Hallucination Benchmarks}

\begin{itemize}
    \item \textbf{Object HalBench}~\cite{rohrbach2018object} is based on MSCOCO \cite{lin2014microsoft} and evaluates object hallucination in image captions.
    Evaluation is restricted to the 80 MSCOCO object categories with segmentation annotations.
    Object presence is determined using both MSCOCO segmentation labels and reference captions, and generated object mentions are normalized through singularization and synonym matching.

    We report two hallucination metrics, Resp. and Ment.
    Resp. measures the fraction of generated responses that contain at least one hallucinated object, while Ment. measures the fraction of object mentions that are hallucinated:
    \begin{equation}
        \text{Resp.}
        =
        \frac{
            N_{\text{resp.}}^{\text{hal.}}
        }{
            N_{\text{resp.}}
        },
        \qquad
        \text{Ment.}
        =
        \frac{
            N_{\text{ment.}}^{\text{hal.}}
        }{
            N_{\text{ment.}}
        }.
        \label{eq:object_halbench}
    \end{equation}
    Here, $N_{\text{resp.}}$ is the total number of generated responses, and $N_{\text{resp.}}^{\text{hal.}}$ is the number of responses containing at least one hallucinated object.
    Similarly, $N_{\text{ment.}}$ is the total number of generated object mentions, and $N_{\text{ment.}}^{\text{hal.}}$ is the number of hallucinated object mentions.
    Lower values indicate less object hallucination.

    \item \textbf{AMBER}~\cite{wang2023amber} contains 1,004 manually selected images annotated with visible objects, attributes, relations, and cognitively plausible hallucination target objects.
    We use its generative evaluation split, where models are prompted with \textit{``Describe this image.''} and hallucination is evaluated from the generated image description.

    Given a response $R$, noun mentions are extracted as candidate objects $R_\text{obj}$ and filtered using the AMBER object vocabulary $X_\text{obj}$:
    \begin{equation}
        R'_\text{obj} = R_\text{obj} \cap X_\text{obj}.
        \label{eq:amber_filtered_objects}
    \end{equation}
    The filtered object mentions $R'_\text{obj}$, which are then compared with the ground-truth object set $A_\text{obj}$, and the hallucination target set $H_\text{obj}$.
    We report three metrics: CHAIR, Hal., and Cog.

    CHAIR measures the fraction of generated object mentions ungrounded in the image:
    \begin{equation}
        \mathrm{CHAIR}(R)
        =
        1
        -
        \frac{
            \left|R'_\text{obj} \cap A_\text{obj}\right|
        }{
            \left|R'_\text{obj}\right|
        }.
        \label{eq:amber_chair}
    \end{equation}

    Hal. is a response-level hallucination indicator that checks whether the response contains at least one ungrounded object:
    \begin{equation}
        \mathrm{Hal}(R)
        =
        \mathbb{I}\left[\mathrm{CHAIR}(R) \neq 0\right].
        \label{eq:amber_hal}
    \end{equation}

    Cog. measures the fraction of generated object mentions that match AMBER's cognitively plausible hallucination targets, i.e., absent objects that are likely to be imagined from the visual context:
    \begin{equation}
        \mathrm{Cog}(R)
        =
        \frac{
            \left|R'_\text{obj} \cap H_\text{obj}\right|
        }{
            \left|R'_\text{obj}\right|
        }.
        \label{eq:amber_cog}
    \end{equation}
    All three metrics are averaged over the generative-evaluation samples, and lower values indicate better performance.

    \item \textbf{HallusionBench}~\cite{guan2024hallusionbench} evaluates image-context reasoning in VLMs, focusing on language hallucination and visual illusion.
    It consists of 346 images and 1,129 human-crafted yes/no questions across diverse visual formats, including charts, tables, maps, OCR images, optical illusions, math figures, posters, and video-like image sequences.

    We report the overall question-level accuracy, denoted as All Acc. (aAcc).
    For each valid image-question pair $(\mathbf{I},\mathbf{q})$, the model prediction is mapped to a binary correctness value $b_M(\mathbf{I},\mathbf{q})\in\{0,1\}$, where 1 indicates a correct prediction, and 0 indicates an incorrect prediction.
    The metric is computed as the average correctness over all valid pairs:
    \begin{equation}
        \mathrm{aAcc}
        =
        \frac{
            \sum_{(\mathbf{I},\mathbf{q})\in \mathcal{V}} b_M(\mathbf{I},\mathbf{q})
        }{
            |\mathcal{V}|
        }.
        \label{eq:hallusionbench_aacc}
    \end{equation}
    Here, $\mathcal{V}$ is the set of valid image-question pairs. Higher values indicate better performance.

\end{itemize}

\subsubsection{General benchmarks}

\begin{itemize}
    \item \textbf{MMStar}~\cite{chen2024we} is a vision-indispensable multimodal benchmark consisting of 1,500 human-curated samples across 6 core capabilities and 18 fine-grained axes.
    It evaluates the visual understanding and reasoning abilities of large vision-language models across coarse perception, fine-grained perception, instance reasoning, logical reasoning, science and technology, and mathematics. The benchmark is designed to reduce text-only shortcuts and data leakage by requiring models to rely on visual content to answer each question.

    \item \textbf{MMVP}~\cite{tong2024eyes} evaluates the fine-grained visual grounding ability of multimodal LLMs through visual question answering on CLIP-blind image pairs.
    These image pairs are visually different to humans but are encoded similarly by CLIP, making them useful for testing whether a model can recognize subtle visual details.
    The benchmark covers basic visual patterns, including orientation and direction, presence of specific features, state and condition, counting, positional relationships, color and appearance, structural characteristics, text, and viewpoint. Following the official evaluation protocol, an image pair is counted as correct only when both associated questions are answered correctly.

    \item \textbf{CV-Bench}~\cite{tong2024cambrian} is a vision-centric benchmark for evaluating the fundamental visual understanding capabilities of multimodal LLMs.
    It reformulates standard computer vision benchmarks into VQA-style questions to assess both 2D and 3D perception. The 2D tasks evaluate spatial relationships and object counting, while the 3D tasks evaluate depth order and relative distance. The benchmark contains 2,638 manually inspected examples constructed from ADE20K \cite{zhou2017scene}, MSCOCO \cite{lin2014microsoft}, and Omni3D \cite{brazil2023omni3d}.
    For CV-Bench, Acc. is computed as the average accuracy over the 2D and 3D task groups.
\end{itemize}

\subsubsection{Vision encoder evaluation}

\begin{itemize}
    \item \textbf{ImageNet-1k}~\cite{imagenet15russakovsky} is a large-scale object recognition benchmark with 1,000 semantic categories. We use ImageNet-1k to evaluate the global semantic representation quality of the vision encoder through linear probing, where the vision encoder is frozen, and only a data-specific linear classifier is trained on top of the extracted visual features. We report classification accuracy (Acc.), where higher values indicate better performance.

    \item \textbf{ImageNet-O}~\cite{hendrycks2021nae} is an out-of-distribution object recognition benchmark designed to evaluate model robustness beyond the ImageNet distribution. We use ImageNet-O to examine whether the learned visual representations remain effective on uncommon and challenging object-centric images. Since ImageNet-O is used in a few-shot evaluation setting, we select a small number of samples from its test set for few-shot training. Specifically, the vision encoder is frozen, and only a data-specific linear classifier is trained using the extracted visual features of the selected few-shot samples. We then evaluate the classifier on the remaining ImageNet-O test samples. This benchmark complements ImageNet-1k by evaluating whether visual features remain effective across diverse distributions. We report classification accuracy (Acc.), where higher values indicate better performance.
    
    \item \textbf{Rendered-SST2}~\cite{socher2013recursive, radford2021learning} is constructed by rendering sentences from the SST-2 sentiment classification dataset as images. We use Rendered-SST2 under the same linear-probing setting, where the vision encoder is frozen, and only a data-specific linear classifier is trained on top of the extracted visual features.
    This benchmark evaluates whether the vision encoder preserves semantic information in text-rendered visual inputs. Unlike object-centric recognition datasets, Rendered-SST2 tests whether visual representations can capture image-based textual semantics.
    We report classification accuracy (Acc.), where higher values indicate better performance.

    \item \textbf{MSCOCO}~\cite{lin2014microsoft} contains images with object-level annotations, including segmentation masks.
    We use MSCOCO for segmentation evaluation to assess whether the vision encoder can distinguish fine-grained object-level information.
    This evaluation complements linear probing by measuring local visual grounding rather than only global image-level semantics.
    We report Recall, where higher values indicate better performance.
\end{itemize}

\begin{algorithm}[t]
\caption{\methodname{}}
\label{alg:vipsy}
\begin{algorithmic}[1]
\Require Dataset $\mathcal{D}=\{(\vx_i,\vm_i)\}_{i=1}^{M}$; policy $\pi_\vtheta$; judge $\pi_{\bm{\phi}}$; T2I model $P_{\mathbf{\psi}}$; number of self-captions/synthetic variants $N_c$; number of cue-conditioned rollouts $N_s$; prompts $\vx_{\text{cap.}}$ for captioning, $\vx_{\text{comp.}}$ for image comparison, $\vx_{\text{agg.}}$ for cue aggregation, $\vx_{\text{cue-cap.}}$ for cue-conditioned rollout, and $\vx_{\text{vis.}}$ for visual-grounding preference selection
\Ensure Preference dataset $\mathcal{D}_{\text{pref}}$

\State $\mathcal{D}_{\text{pref}} \leftarrow \emptyset$
\Comment{Initialize preference dataset}

\For{each $(\vx_i,\vm_i)\in\mathcal{D}$}

\Statex \textbf{\{Stage 1: Self-captioned visual cue extraction\}}

\State $\{\vq_j\}_{j=1}^{N_c} \sim \pi_{\vtheta}(\cdot \mid \vx_{\text{cap.}}, \vm_i)$
\Comment{Generate $N_c$ self-captions}

\State $\{\vm'_{i,j}\}_{j=1}^{N_c} = \{P_{\mathbf{\psi}}(\vq_j)\}_{j=1}^{N_c}$
\Comment{Synthesize semantic image variants}

\State $\vc_{i,j} = \pi_{\bm{\phi}}(\vx_{\text{comp.}}, \vm_i, \vm'_{i,j})$
\Comment{Extract consistent visual cues}

\State $\vc_i = \pi_{\bm{\phi}}(\vx_{\text{agg.}}, \{\vc_{i,j}\}_{j=1}^{N_c})$
\Comment{Aggregate consistent visual cues}

\Statex \textbf{\{Stage 2: Cue-conditioned preference synthesis\}}

\State $\vy_i^{(s)} \sim \pi_{\vtheta}(\cdot \mid \vx_{\text{cue-cap.}}, \vm_i, \vc_i), \quad s=1,\ldots,N_s$
\Comment{Sample cue-conditioned responses}

\State $(\vy_{w,i}, \vy_{l,i}) =
\pi_{\bm{\phi}}\!\left(\vx_{\text{vis.}}, \vm_i, \{\vy_i^{(s)}\}_{s=1}^{N_s}\right)$
\Comment{Select visually grounded and less grounded responses}

\State $\mathcal{D}_{\text{pref}}
\leftarrow
\mathcal{D}_{\text{pref}}
\cup
\{(\vx_i,\vm_i,\vy_{w,i},\vy_{l,i})\}$
\Comment{Store preference pair}

\EndFor

\State \Return $\mathcal{D}_{\text{pref}}$
\end{algorithmic}
\end{algorithm}

\section{Algorithm}
\label{ap:algorithm}

Algorithm~\ref{alg:vipsy} presents the ViPSy preference construction pipeline. 
For each image-prompt pair, ViPSy first generates multiple self-captions using the policy model and synthesizes semantic image variants with a text-to-image model. 
A judge model compares each variant with the original image to extract shared object-level evidence, and aggregates these pairwise cues into a consistent visual cue. 
Then, the policy model samples multiple responses conditioned on the original prompt, image, and visual cue. 
The judge selects the most visually grounded candidate as the preferred response and the least grounded candidate as the dispreferred response, forming a preference pair. 
The synthetic images and visual cues are used only for constructing preference data and are not required during inference.

\section{Additional results}
\label{ap:additional}

We provide the full results discussed in Section~\ref{sec:experiments}. Specifically, Hal. AVG. denotes
the average score over Object HalBench and AMBER, which measure hallucination in generated responses.
Gen. AVG. denotes the average score over MMStar, MMVP, and CV-Bench, which are used to evaluate general
visual-language capabilities in Section~\ref{sec:experiments}.

\subsection{Policy model ablation}
\label{ap:policy_abl}
Table~\ref{tab:ap_policy_ablation} reports the results of applying preference alignment to different policy models using policy-specific preference data constructed by \methodname{}. The results show that \methodname{} not only
mitigates hallucination but also improves general vision-language capabilities. This indicates that \methodname{} is not limited to a single policy model and can generalize across
different VLMs.

\begin{table*}[t]
\centering
\caption{Results on hallucination and general benchmarks using various policy models. We apply \methodname{} to diverse policy models.}
\setlength{\tabcolsep}{3.5pt}
\begin{adjustbox}{width=\textwidth}
\begin{tabular}{@{}l|cc ccc c|ccc@{}}
\toprule

& \multicolumn{6}{c|}{\textbf{Hallucination benchmarks}}
& \multicolumn{3}{c}{\textbf{General benchmarks}} \\

\cmidrule(lr){2-7}
\cmidrule(lr){8-10}

\textbf{Method}
& \multicolumn{2}{c}{Object HalBench \cite{rohrbach2018object}}
& \multicolumn{3}{c}{AMBER \cite{wang2023amber}}
& Hal. Bench \cite{guan2024hallusionbench}
& MMStar \cite{chen2024we}
& MMVP \cite{tong2024eyes}
& CV-Bench \cite{tong2024cambrian} \\

\cmidrule(lr){2-3}
\cmidrule(lr){4-6}
\cmidrule(lr){7-7}
\cmidrule(lr){8-8}
\cmidrule(lr){9-9}
\cmidrule(lr){10-10}

& Resp.$\downarrow$ & Ment.$\downarrow$
& CHAIR$\downarrow$ & Hal.$\downarrow$ & Cog.$\downarrow$
& aAcc.$\uparrow$
& Acc.$\uparrow$ & Acc.$\uparrow$ & Acc.$\uparrow$ \\

\midrule

\rowcolor{gray!15}
\multicolumn{10}{@{}c@{}}{\textbf{Qwen2.5-VL-7B}} \\
\midrule
Vanilla

& 9.0     & 5.3   & 5.0     & 26.4  & 1.7   & 57.7  & 59.8  & 56.6  & 78.3 \\
\rowcolor{cyan!15}
ViPSy (\textbf{Ours})

&7.6   & 4.8   & 4.1   & 23.3  & 1.0     & 60.4  & 61.7  & 58.6  & 78.0 \\

\midrule

\rowcolor{gray!15}
\multicolumn{10}{@{}c@{}}{\textbf{mPLUG-Owl3-7B}} \\
\midrule
Vanilla

& 11.6  & 6.6   & 5.4   & 8.3   & 1.4   & 50.8  & 48.3  & 43.3  & 55.0 \\

\rowcolor{cyan!15}
ViPSy (\textbf{Ours})

& 9.6   & 5.9   & 4.8   & 7.8   & 1.1   & 54.6  & 48.2  & 45.3  & 56.8 \\

\midrule

\rowcolor{gray!15}
\multicolumn{10}{@{}c@{}}{\textbf{InternVL3.5-8B}} \\
\midrule
Vanilla

& 8.3   & 4.6   & 4.3   & 24.6  & 1.5   & 64.2  & 66.3  & 59.3  & 81.6 \\

\rowcolor{cyan!15}
ViPSy (\textbf{Ours})

& 7.3   & 4.2   & 3.9   & 17.3  & 1.2   & 65.4  & 67.9  & 60.0    & 81.4 \\
\midrule

\rowcolor{gray!15}
\multicolumn{10}{@{}c@{}}{\textbf{Kimi-VL-A3B}} \\
\midrule
Vanilla

& 5.6   & 3.5   & 5.9   & 38.8  & 3.8   & 56.0    & 49.5  & 48.6  & 75.8 \\

\rowcolor{cyan!15}
ViPSy (\textbf{Ours})

& 5.0     & 3.1   & 4.1   & 29.4  & 1.9   & 57.7  & 52.2  & 50.0    & 76.4 \\

\bottomrule
\end{tabular}

\end{adjustbox}
\label{tab:ap_policy_ablation}
\end{table*}

\subsection{Model component ablation}
\label{ap:model_abl}
In Table~\ref{tab:ap_model_ablation}, we analyze how each model component in \methodname{} affects performance. We vary the image-to-text captioning model, the text-to-image synthesis model, and the
rollout model used in \methodname{}. The comparison highlights that self-captioning and self-rollouts are
important for keeping the constructed preference data close to the policy distribution, while the choice
of text-to-image model has a smaller effect.

\begin{table*}[t]
\centering
\setlength{\tabcolsep}{3.5pt}
\caption{Results on hallucination and general benchmarks using LLaVA-1.5-7B. We vary the image-to-text captioning model (I$\rightarrow$T), the text-to-image synthesis model (T$\rightarrow$I), and the rollout model while keeping the remaining components fixed. ``Self'' denotes using the policy model itself.}
\begin{adjustbox}{width=\textwidth}
\begin{tabular}{@{}lll|cc ccc c|ccc@{}}
\toprule

\multicolumn{3}{@{}c|}{\textbf{\methodname{} Configuration}}
& \multicolumn{6}{c|}{\textbf{Hallucination benchmarks}}
& \multicolumn{3}{c}{\textbf{General benchmarks}} \\

\cmidrule(lr){1-3}
\cmidrule(lr){4-9}
\cmidrule(lr){10-12}

\textbf{I$\rightarrow$T}
& \textbf{T$\rightarrow$I}
& \textbf{Rollout}
& \multicolumn{2}{c}{Object HalBench }
& \multicolumn{3}{c}{AMBER }
& Hal. Bench 
& MMStar 
& MMVP 
& CV-Bench  \\

\cmidrule(lr){4-5}
\cmidrule(lr){6-8}
\cmidrule(lr){9-9}
\cmidrule(lr){10-10}
\cmidrule(lr){11-11}
\cmidrule(lr){12-12}

& &
& Resp.$\downarrow$ & Ment.$\downarrow$
& CHAIR$\downarrow$ & Hal.$\downarrow$ & Cog.$\downarrow$
& aAcc.$\uparrow$
& Acc.$\uparrow$ & Acc.$\uparrow$ & Acc.$\uparrow$ \\

\midrule

\rowcolor{red!10}
\multicolumn{3}{@{}l|}{\textit{Vanilla (LLaVA-1.5-7B)}}
& 51.3 & 25.9
& 7.9 & 36.8 & 4.3
& 46.8
& 33.4 & 23.3 & 61.9 \\
\midrule

\rowcolor{cyan!15}
Self
& SD-3.5-L 
& Self
& {4.0}& {2.6}& {1.8}& {9.6}& {0.6}& {48.1}& {35.4}& {30.6}& {64.4} \\
\midrule

GPT-5 mini
& \textcolor{lightgray}{SD-3.5-L}
& \textcolor{lightgray}{Self}
& 6.6   & 4.4   & 1.9   & 12.0    & 0.7   &  47.2     & 33.8  & 29.3  & 64.6 \\

Qwen3-VL-30B
& \textcolor{lightgray}{SD-3.5-L}
& \textcolor{lightgray}{Self}
& 4.6   & 3.5   & 2.2   & 12.4  & 0.8   & 47.9      & 34.4  & 28.0    & 64.0 \\

Qwen2.5-VL-72B
& \textcolor{lightgray}{SD-3.5-L}
& \textcolor{lightgray}{Self}
& 4.0     & 2.2   & 2.1   & 12.1  & 0.9   &    47.7   & 34.8  & 29.3  & 63.1 \\

\midrule

\textcolor{lightgray}{Self}
& PixelDiT 
& \textcolor{lightgray}{Self}
& 3.3   & 2.4   & 2.4   & 13.5  & 0.9   &  48.7     & 34.7  & 28.6  & 64.8 \\

\textcolor{lightgray}{Self}
& Flux-1.0-dev
& \textcolor{lightgray}{Self}
& 4.6   & 3.1   & 2.3   & 11.1  & 0.7   &  48.7     & 34.6  & 29.3  & 63.6 \\
\textcolor{lightgray}{Self}
& SD-3.5-M 
& \textcolor{lightgray}{Self}
& 4.3   & 2.6   & 1.7   & 10.1  & 0.6   &  48.2     & 34.6  & 29.3  & 64.7 \\
\midrule

\textcolor{lightgray}{Self}
& \textcolor{lightgray}{SD-3.5-L}
& GPT-5 mini
& 30.6  & 15.1  & 5.3   & 24.3  & 2.9   &    47.7   & 33.0    & 28.0    & 62.2 \\

\textcolor{lightgray}{Self}
& \textcolor{lightgray}{SD-3.5-L}
& Qwen3-VL-30B
& 29.6  & 16.6  & 5.1   & 25.6  & 3.1   &    47.9   & 33.1  & 30.6  & 62.5 \\
\textcolor{lightgray}{Self}
& \textcolor{lightgray}{SD-3.5-L}
& Qwen2.5-VL-72B
& 21.6  & 10.4  & 3.8   & 18.5  & 1.8   &    47.7   & 33.7  & 28.6  & 63.1 \\

\bottomrule
\end{tabular}

\end{adjustbox}
\label{tab:ap_model_ablation}
\end{table*}

\subsection{Number of rollouts ablation}
\label{ap:num_rollout_ablation}
Table~\ref{tab:ap_num_roll_ablation} reports the results for different numbers of cue-conditioned rollouts. Overall, increasing the number of rollouts generally improves performance across both hallucination and general benchmarks, as a larger candidate pool provides more diverse responses for preference selection. In particular, using more rollouts enables the judge to compare candidates with different grounding quality under the same image, prompt, and visual cue, resulting in more informative preference pairs for training. While 16 rollouts further improve several metrics, the gains over 8 rollouts are relatively marginal, and 8 rollouts even achieve better performance on CV-Bench. Considering the additional computational cost incurred by generating more rollouts, we choose 8 rollouts as a practical default.

\begin{table*}[t]
\centering
\caption{Results on hallucination and general benchmarks using LLaVA-1.5-7B. We compare preference construction with different numbers of cue-conditioned rollouts. For each rollout setting, we report the mean and standard deviation over five runs.}
\setlength{\tabcolsep}{3.5pt}
\begin{adjustbox}{width=\textwidth}
\begin{tabular}{@{}c|cc ccc c|ccc@{}}
\toprule

& \multicolumn{6}{c|}{\textbf{Hallucination benchmarks}}
& \multicolumn{3}{c}{\textbf{General benchmarks}} \\

\cmidrule(lr){2-7}
\cmidrule(lr){8-10}

\shortstack[c]{\textbf{Number of}\\\textbf{Rollouts}}
& \multicolumn{2}{c}{Object HalBench \cite{rohrbach2018object}}
& \multicolumn{3}{c}{AMBER \cite{wang2023amber}}
& Hal. Bench \cite{guan2024hallusionbench}
& MMStar \cite{chen2024we}
& MMVP \cite{tong2024eyes}
& CV-Bench \cite{tong2024cambrian} \\

\cmidrule(lr){2-3}
\cmidrule(lr){4-6}
\cmidrule(lr){7-7}
\cmidrule(lr){8-8}
\cmidrule(lr){9-9}
\cmidrule(lr){10-10}

& Resp.$\downarrow$ & Ment.$\downarrow$
& CHAIR$\downarrow$ & Hal.$\downarrow$ & Cog.$\downarrow$
& aAcc.$\uparrow$
& Acc.$\uparrow$ & Acc.$\uparrow$ & Acc.$\uparrow$ \\

\midrule

\rowcolor{red!10}
Vanilla

& $51.3$ & $25.9$
& $7.9$ & $36.8$ & $4.3$
& $46.8$
& $33.4$ & $23.3$ & $61.9$ \\

\midrule

\rowcolor{cyan!15}
$8$

& $3.67 \pm0.53$ & $2.54 \pm0.26 $& $1.92 \pm0.29$ & $9.68 \pm1.52 $&$ 0.64 \pm0.13 $& $48.25  \pm0.67 $&$ 35.39 \pm0.67 $& $ 30.53 \pm1.85$ & $64.27 \pm0.37 $\\

\midrule

$2$

& $4.73 \pm0.55$ & $2.84 \pm0.40$ & $2.82 \pm0.34$ & $13.16 \pm1.66 $& $0.78 \pm0.13 $& $47.65\pm0.61$ & $34.00 \pm0.21$ & $29.27 \pm1.21$ & $63.06 \pm0.62$ \\
$4$

&$ 4.33 \pm0.91$ &$ 2.86 \pm0.53 $& $2.38 \pm0.19 $& $11.90 \pm1.23 $& $0.72 \pm0.13$ & $47.55 \pm0.52$ & $34.59 \pm0.25 $& $29.47 \pm1.66$ & $63.79 \pm0.32 $\\

$16$
& $3.33 \pm0.33$ & $1.84 \pm0.10 $& $2.08 \pm0.04$ &$ 9.40 \pm0.40$ &$ 0.64 \pm0.05 $& $48.83 \pm0.20 $& $35.41 \pm0.10 $&$ 30.73 \pm0.60$ & $64.02 \pm0.19$ \\

\bottomrule
\end{tabular}

\end{adjustbox}
\label{tab:ap_num_roll_ablation}
\end{table*}

\subsection{Preference selection judge ablation}
\label{ap:selection_judge_ablation}
Table~\ref{tab:ap_judge_ablation} provides the complete benchmark results under different preference selection judge models. Since the judge selects preferred and dispreferred responses from multiple candidates, this ablation examines whether \methodname{} depends on a particular judge model. The results remain stable across judges, indicating that the preference selection process is robust to the choice of judge model.

\begin{table*}[t]
\centering
\caption{Results on hallucination and general benchmarks using LLaVA-1.5-7B. We compare preference construction by varying the preference selection judge in \methodname{}.}
\setlength{\tabcolsep}{3.5pt}
\begin{adjustbox}{width=\textwidth}
\begin{tabular}{@{}l|cc ccc c|ccc@{}}
\toprule

& \multicolumn{6}{c|}{\textbf{Hallucination benchmarks}}
& \multicolumn{3}{c}{\textbf{General benchmarks}} \\

\cmidrule(lr){2-7}
\cmidrule(lr){8-10}

\textbf{Judge}
& \multicolumn{2}{c}{Object HalBench \cite{rohrbach2018object}}
& \multicolumn{3}{c}{AMBER \cite{wang2023amber}}
& Hal. Bench \cite{guan2024hallusionbench}
& MMStar \cite{chen2024we}
& MMVP \cite{tong2024eyes}
& CV-Bench \cite{tong2024cambrian} \\

\cmidrule(lr){2-3}
\cmidrule(lr){4-6}
\cmidrule(lr){7-7}
\cmidrule(lr){8-8}
\cmidrule(lr){9-9}
\cmidrule(lr){10-10}

& Resp.$\downarrow$ & Ment.$\downarrow$
& CHAIR$\downarrow$ & Hal.$\downarrow$ & Cog.$\downarrow$
& aAcc.$\uparrow$
& Acc.$\uparrow$ & Acc.$\uparrow$ & Acc.$\uparrow$ \\

\midrule

GPT-5 mini
& 3.3   & 2.2   & 1.9   & 9.6   & 0.5   & 48.1  & 34.5  & 27.3  & 64.7 \\

Qwen3-VL-30B
& 3.3   & 2.2   & 2.0     & 9.8   & 0.7   & 47.7  & 35.0    & 28.0    & 64.4 \\

\midrule
\rowcolor{cyan!15}
Qwen2.5-VL-72B

& {4.0}& {2.6}& {1.8}& {9.6}& {0.6}& {48.1}& {35.4}& {30.6}& {64.4} \\
\bottomrule
\end{tabular}

\end{adjustbox}
\label{tab:ap_judge_ablation}
\end{table*}

\subsection{Synthesis strategy ablation}
\label{ap:synthesis_strategy}
Table~\ref{tab:ap_synthetic_strategy_ablation} reports results for cue conditioning and visual variation strategies used for cue extraction. The no-cue setting still benefits from policy self-sampling, but it is less effective than cue-conditioned preference construction, indicating that
grounded visual cues are important for guiding rollouts toward visually faithful responses. For visual variation, simple transformations such as translation, resizing, masking, and cropping improve performance but provide limited semantic diversity. Video-based variation uses an image-to-video model~\cite{wan2.1} to generate candidate frames from the input image. We then compute the cosine similarity between the visual features of the original image and each generated frame, and sample frames whose similarity is closest to 0.8. This selection encourages moderate visual variation while maintaining semantic consistency with the original image. However, despite this filtering, uncontrolled changes across generated frames can still make the extracted cues less reliable. In contrast, diffusion-based semantic synthesis produces richer semantic variations and achieves the strongest overall results, showing that semantically diverse visual variations help extract more consistent and reliable object-level cues.

\begin{table*}[t]
\centering
\caption{Results on hallucination and general benchmarks using LLaVA-1.5-7B. We compare preference construction without visual cues, simple image transformations (translation, resize, mask, and crop), video-based variation, and diffusion-based semantic synthesis.}
\setlength{\tabcolsep}{3.5pt}
\begin{adjustbox}{width=\textwidth}
\begin{tabular}{@{}l|cc ccc c|ccc@{}}
\toprule

& \multicolumn{6}{c|}{\textbf{Hallucination benchmarks}}
& \multicolumn{3}{c}{\textbf{General benchmarks}} \\

\cmidrule(lr){2-7}
\cmidrule(lr){8-10}

\textbf{Strategy}
& \multicolumn{2}{c}{Object HalBench \cite{rohrbach2018object}}
& \multicolumn{3}{c}{AMBER \cite{wang2023amber}}
& Hal. Bench \cite{guan2024hallusionbench}
& MMStar \cite{chen2024we}
& MMVP \cite{tong2024eyes}
& CV-Bench \cite{tong2024cambrian} \\

\cmidrule(lr){2-3}
\cmidrule(lr){4-6}
\cmidrule(lr){7-7}
\cmidrule(lr){8-8}
\cmidrule(lr){9-9}
\cmidrule(lr){10-10}

& Resp.$\downarrow$ & Ment.$\downarrow$
& CHAIR$\downarrow$ & Hal.$\downarrow$ & Cog.$\downarrow$
& aAcc.$\uparrow$
& Acc.$\uparrow$ & Acc.$\uparrow$ & Acc.$\uparrow$ \\

\midrule

\rowcolor{red!10}
Vanilla

& 51.3 & 25.9
& 7.9 & 36.8 & 4.3
& 46.8
& 33.4 & 23.3 & 61.9 \\

\midrule

w/o cue

& 7.6   & 3.9   & 3.3   & 19.7  & 1.2   &   47.1    & 34.3  & 28.6  & 64.6 \\
\midrule

Translation

& 5.3   & 4.6   & 2.3   & 12.5  & 0.8   &    48.6   & 34.3  & 30.6  & 64.5 \\
Resize

& 4.3   & 3.2   & 2.4   & 13.4  & 0.8   &    47.2   & 35.0    & 26.6  & 65.8 \\
Mask

& 5.0     & 3.7   & 2.4   & 12.5  & 0.5   &   47.2    & 34.9  & 28.0     & 64.7 \\

Crop

& 5.0     & 3.9   & 2.3   & 11.8  & 0.8   &   47.6    & 34.6  & 28.0     & 64.7 \\

Video

& 4.3   & 3.4   & 2.3   & 11.5  & 0.8   &    48.0   & 33.2  & 28.0     & 63.9 \\
\midrule

\rowcolor{cyan!15}
Diffusion

& {4.0}& {2.6}& {1.8}& {9.6}& {0.6}& {48.1}& {35.4}& {30.6}& {64.4} \\

\bottomrule
\end{tabular}

\end{adjustbox}
\label{tab:ap_synthetic_strategy_ablation}
\end{table*}

\subsection{Variance of the \methodname{}}
\label{ap:infer_variance}

In the main experiments, we follow prior work and use temperature 0 during inference, which removes randomness in generation and ensures a fair comparison with existing hallucination-mitigation methods. Since VLM outputs can vary when using a nonzero temperature, we further examine the inference-time variance of the model preference-aligned with \methodname{}-constructed data. Specifically, we set the temperature to 0.7 and repeat inference five times. As shown in Table~\ref{tab:ap_infer_var}, which summarizes the mean and standard deviation across the five runs, the model preference-aligned with \methodname{}-constructed data achieves lower mean hallucination scores than the vanilla model on Object HalBench and AMBER, while achieving higher mean scores on HallusionBench and general vision-language benchmarks. Overall, these results show that preference alignment with \methodname{}-constructed data maintains its hallucination-mitigation effect while enhancing general vision-language capability under repeated inference with temperature 0.7. 

\begin{table*}[t]
\centering
\caption{Variance of \methodname{} on LLaVA-1.5-7B. Results are computed over five inference runs with temperature 0.7 and reported as mean and standard deviation.}
\setlength{\tabcolsep}{3.5pt}
\begin{adjustbox}{width=\textwidth}
\begin{tabular}{@{}l|cc ccc c|ccc@{}}
\toprule

& \multicolumn{6}{c|}{\textbf{Hallucination benchmarks}}
& \multicolumn{3}{c}{\textbf{General benchmarks}} \\

\cmidrule(lr){2-7}
\cmidrule(lr){8-10}

\textbf{Method}
& \multicolumn{2}{c}{Object HalBench \cite{rohrbach2018object}}
& \multicolumn{3}{c}{AMBER \cite{wang2023amber}}
& Hal. Bench \cite{guan2024hallusionbench}
& MMStar \cite{chen2024we}
& MMVP \cite{tong2024eyes}
& CV-Bench \cite{tong2024cambrian} \\

\cmidrule(lr){2-3}
\cmidrule(lr){4-6}
\cmidrule(lr){7-7}
\cmidrule(lr){8-8}
\cmidrule(lr){9-9}
\cmidrule(lr){10-10}

& Resp.$\downarrow$ & Ment.$\downarrow$
& CHAIR$\downarrow$ & Hal.$\downarrow$ & Cog.$\downarrow$
& aAcc.$\uparrow$
& Acc.$\uparrow$ & Acc.$\uparrow$ & Acc.$\uparrow$ \\

\rowcolor{red!10}
Vanilla 
& $53.0 \pm 1.9$ & $27.7 \pm 1.1$ & $9.3 \pm 0.8$ & $41.7 \pm 3.0$ & $4.3 \pm 0.2$ & $46.4 \pm 0.3$ & $33.0 \pm 0.8$ & $27.0 \pm 2.4$ & $57.8 \pm 2.4$ \\
\midrule

\rowcolor{cyan!15}
ViPSy (\textbf{Ours})
& $4.1 \pm 0.4$ & $2.8 \pm 0.5$
& $2.0 \pm 0.3$ & $10.2 \pm 1.3$ & $0.6 \pm 0.1$
& $47.8 \pm 0.8$
& $34.4 \pm 0.9$ & $29.8 \pm 1.1$ & $64.6 \pm 1.4$ \\
\bottomrule
\end{tabular}

\end{adjustbox}
\label{tab:ap_infer_var}
\end{table*}

\section{Qualitative results}
\label{ap:addi_qual}

We provide additional qualitative comparisons to illustrate how \methodname{} constructs preference data and how the aligned model behaves after training. We also provide examples of our proposed self-captioned semantic synthesis pipeline,
which visualizes how synthetic image variations are generated from policy-generated captions. 

\subsection{Preference data examples}
\label{ap:preference_data_examples}
\begin{figure*}[t]
  \centering

  \includegraphics[width=\linewidth]{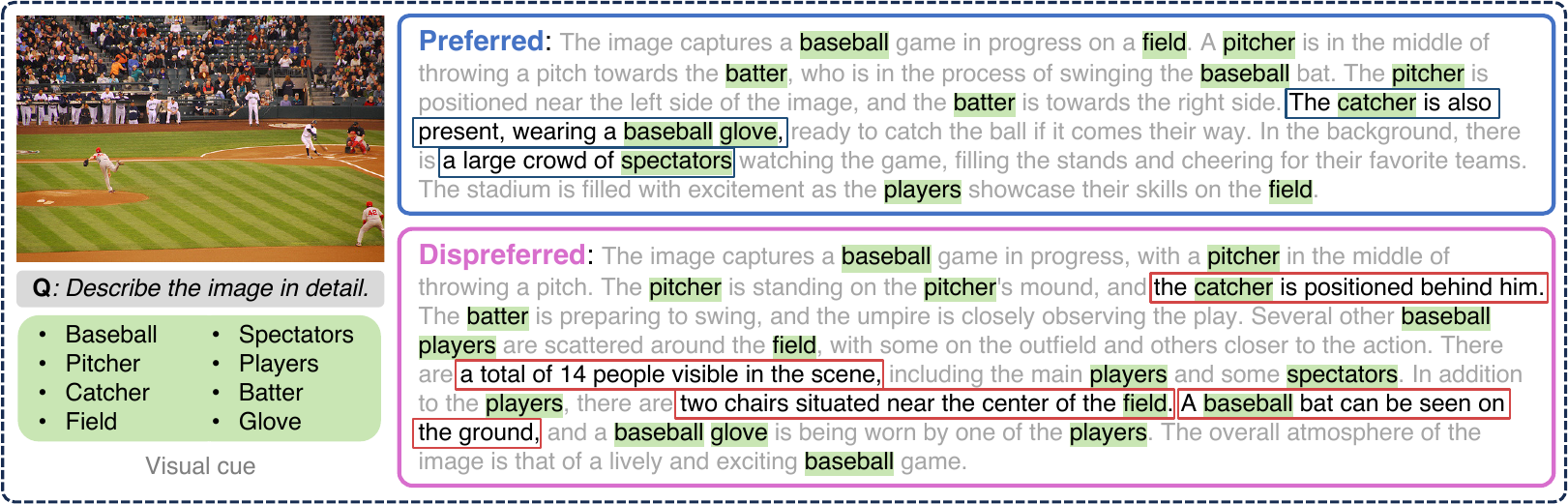}

  \includegraphics[width=\linewidth]{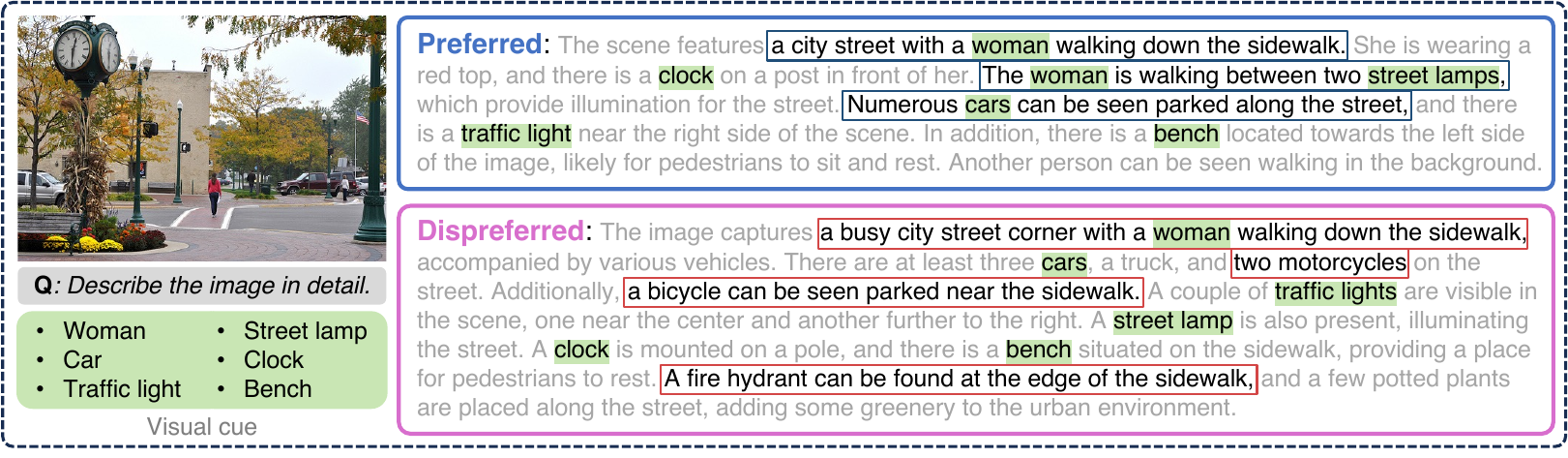}

  \includegraphics[width=\linewidth]{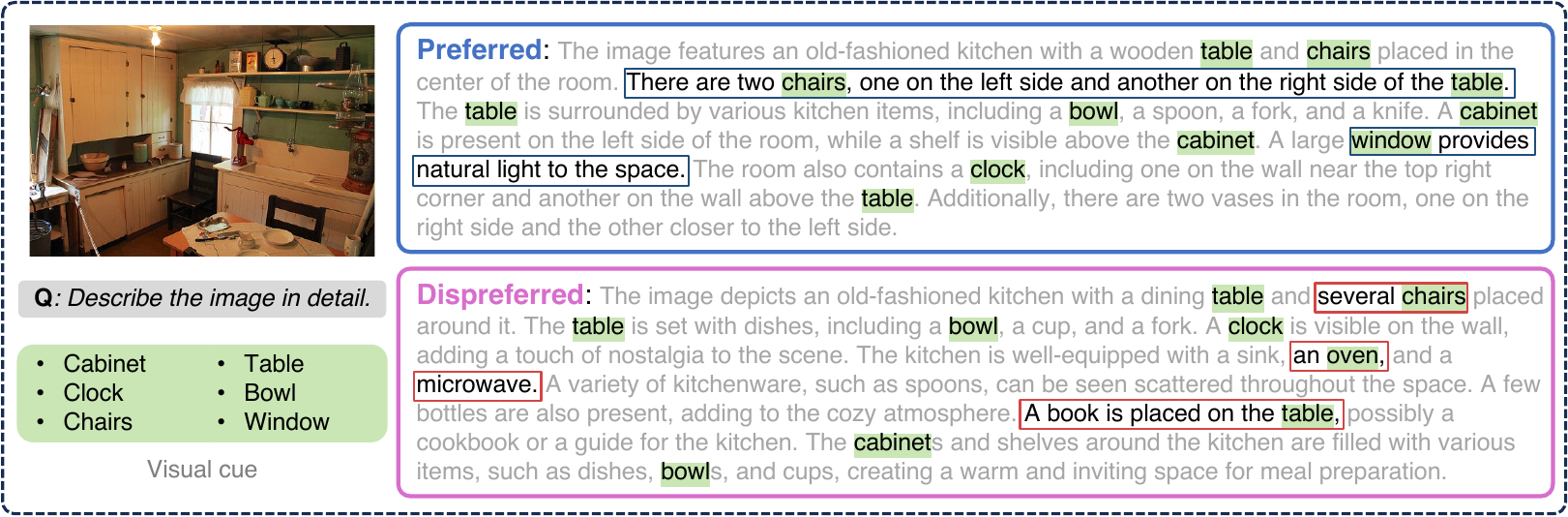}

\caption{Qualitative examples of cue-conditioned preference pairs. Given the same image, prompt, and visual cue, preferred responses remain image-grounded, while dispreferred ones hallucinate.}
  \label{fig:ap_qual_train}
\end{figure*}

Figure~\ref{fig:ap_qual_train} shows additional examples of preference pairs constructed by \methodname{}. Each example contains the image, the prompt, the extracted visual cue, and the selected
preferred and dispreferred responses. The preferred responses describe objects and relations supported by the image and visual cue, whereas the dispreferred responses include information that is not present
in the image. These examples show that cue-conditioned self-rollouts produce candidate responses with different degrees of visual grounding, enabling the judge to select informative preference pairs for preference alignment.

\subsection{Image description examples}
\label{ap:mitigate_hal_qual}
\begin{figure*}[ht]
  \centering

  \includegraphics[width=\linewidth]{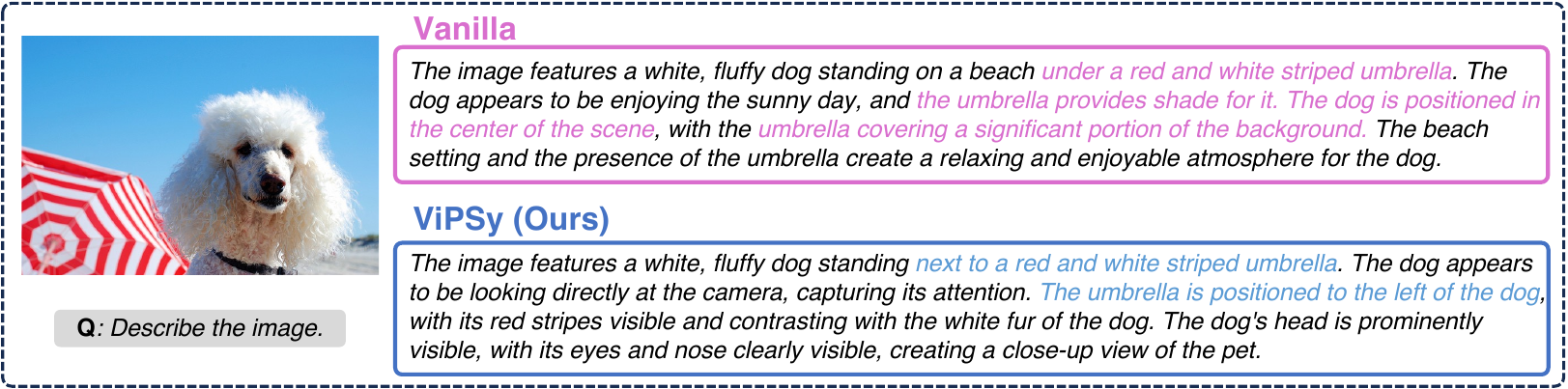}

  \includegraphics[width=\linewidth]{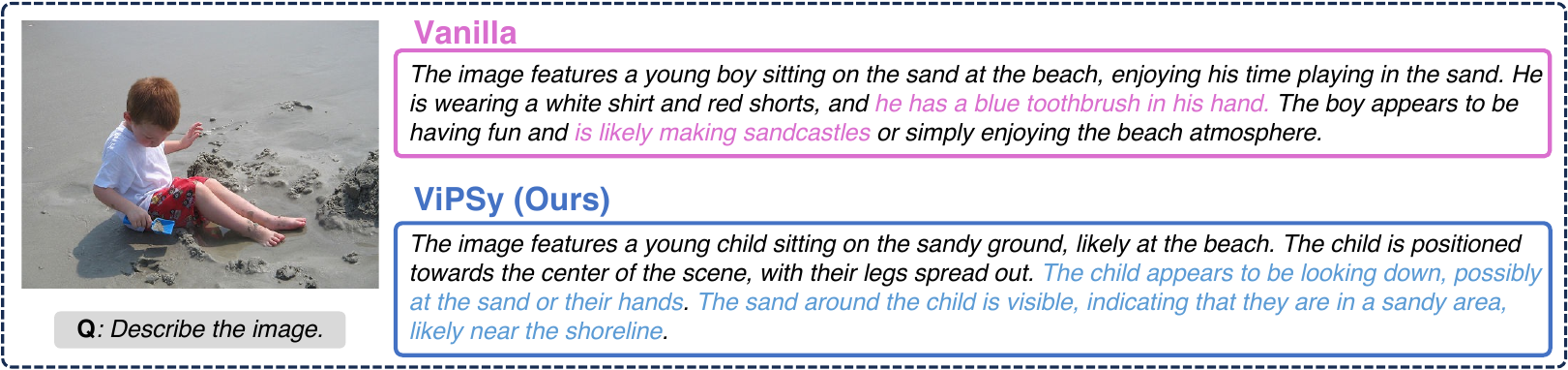}

  \includegraphics[width=\linewidth]{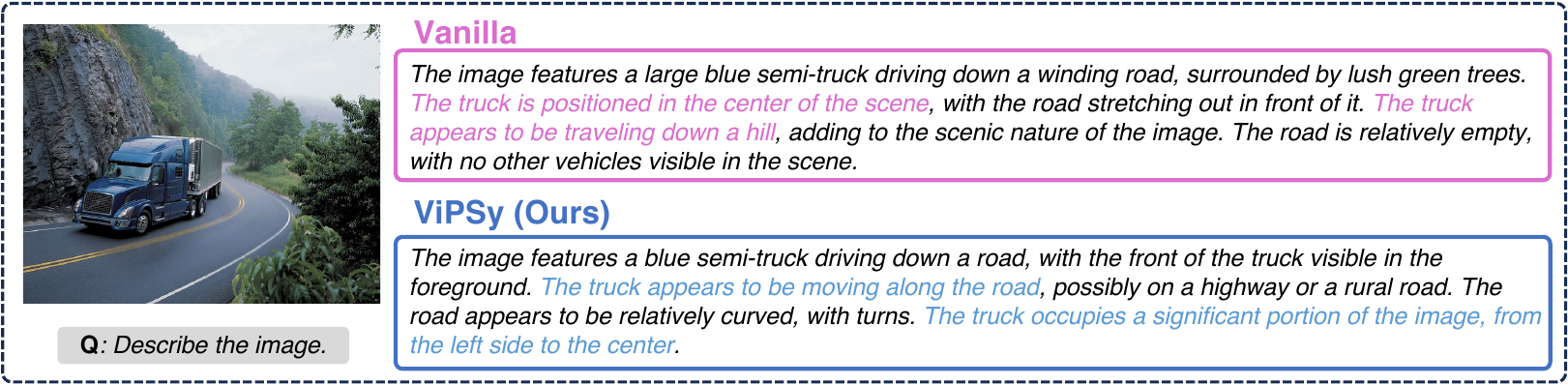}

  \includegraphics[width=\linewidth]{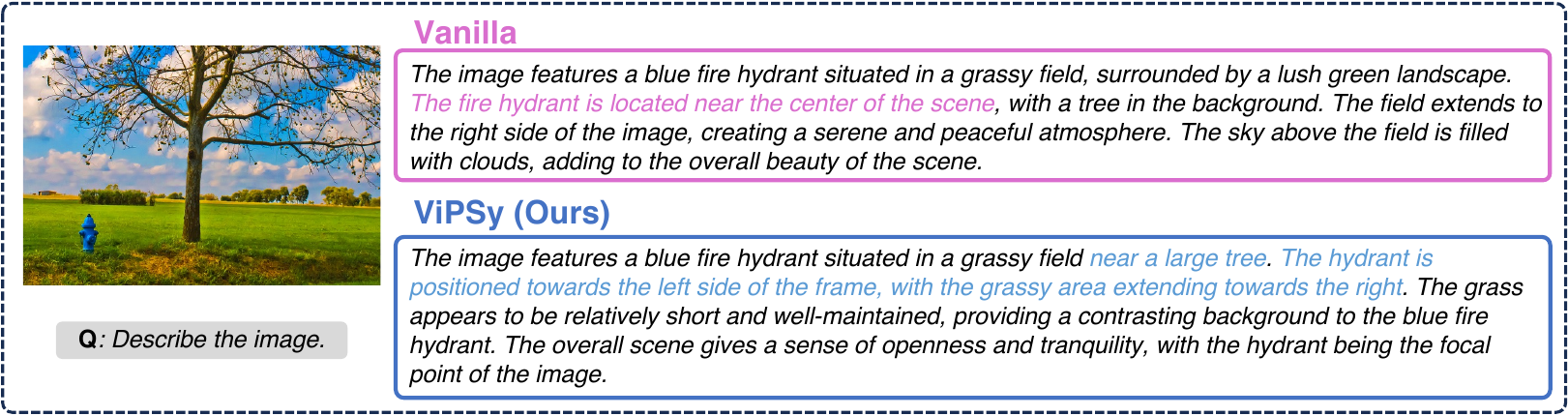}

\caption{Qualitative comparison between the vanilla model and the model preference-aligned using preference pairs constructed by \methodname{}. After preference alignment, the aligned model produces image-grounded descriptions, whereas the vanilla model describes content that is not present in the image. Highlights indicate image-inconsistent descriptions for the vanilla model and grounded descriptions for the preference-aligned model.}
      
  \label{fig:ap_qual_results}
  \vspace{-0.1in}
\end{figure*}

Figure~\ref{fig:ap_qual_results} compares image descriptions from the vanilla model and the model preference-aligned using preference pairs constructed by \methodname{}. The vanilla model often includes hallucinated details that are not supported by the image. In contrast, the model trained with \methodname{} produces descriptions that are better grounded in the image. This supports the effectiveness of the proposed preference construction in
reducing hallucination.

\subsection{Self-captioned semantic synthesis examples}
\label{ap:self-cap-synthesis_qual}

\begin{figure*}[ht]
  \centering

  \includegraphics[width=\linewidth]{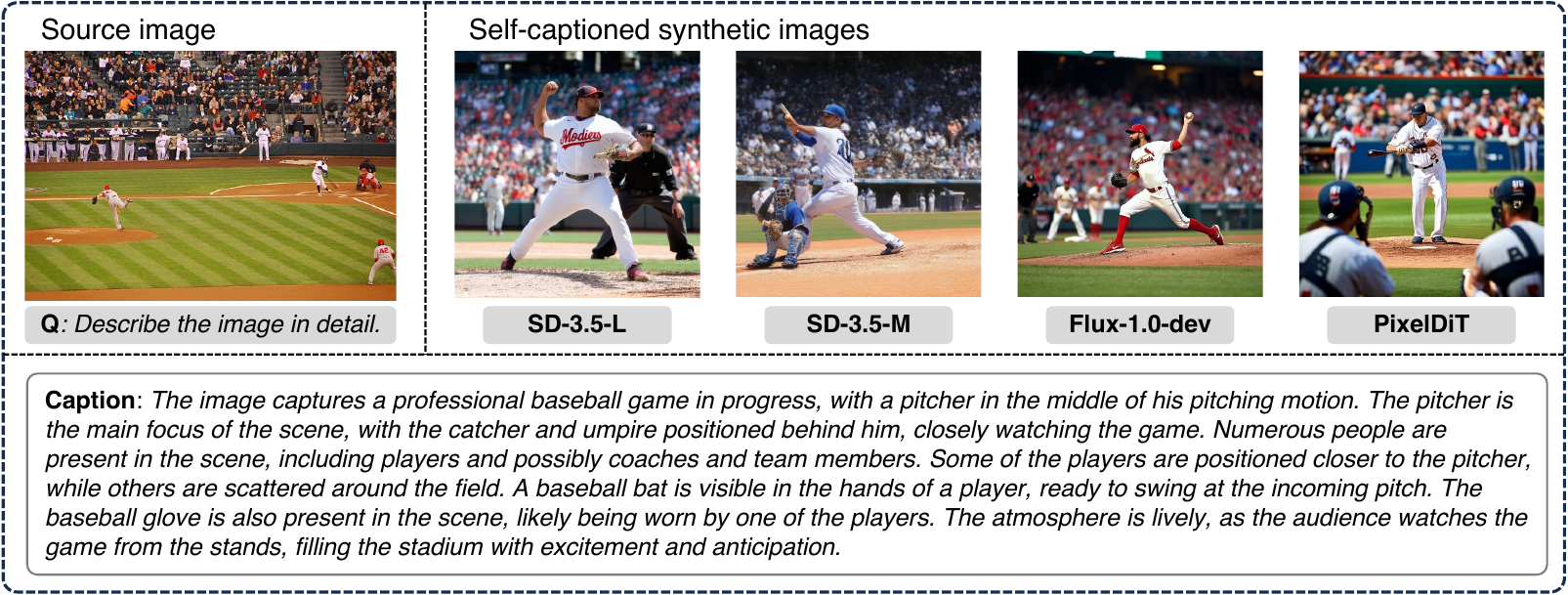}

  \includegraphics[width=\linewidth]{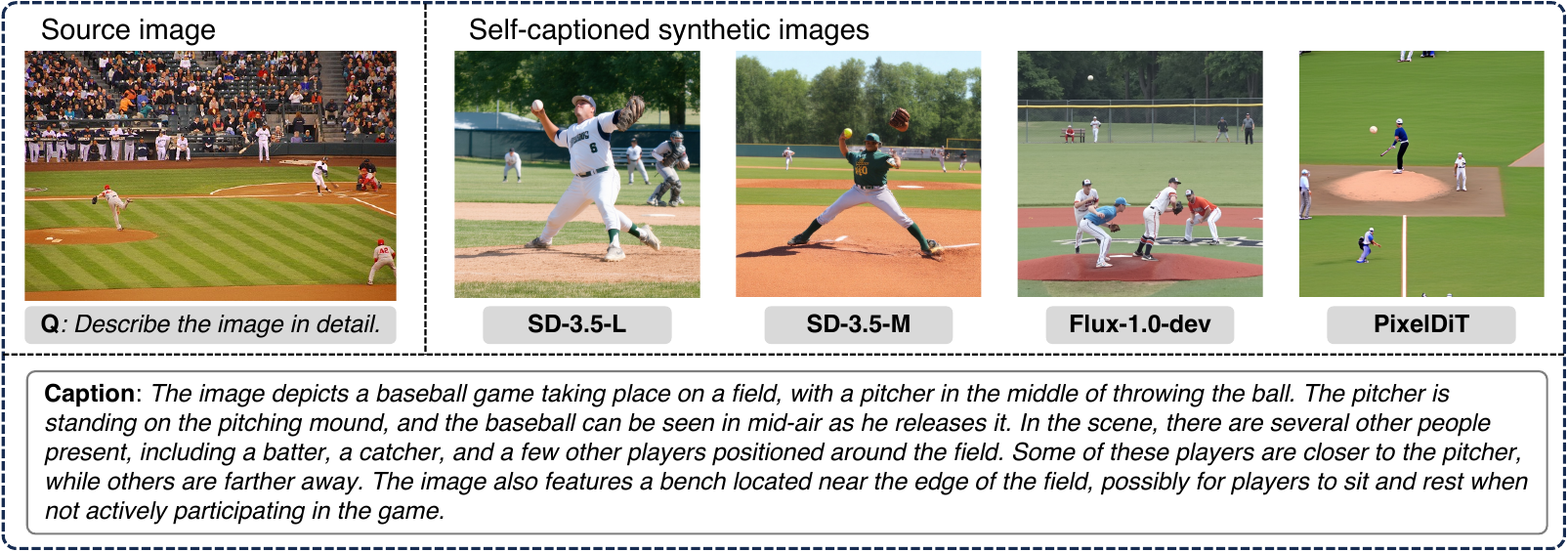}

  \includegraphics[width=\linewidth]{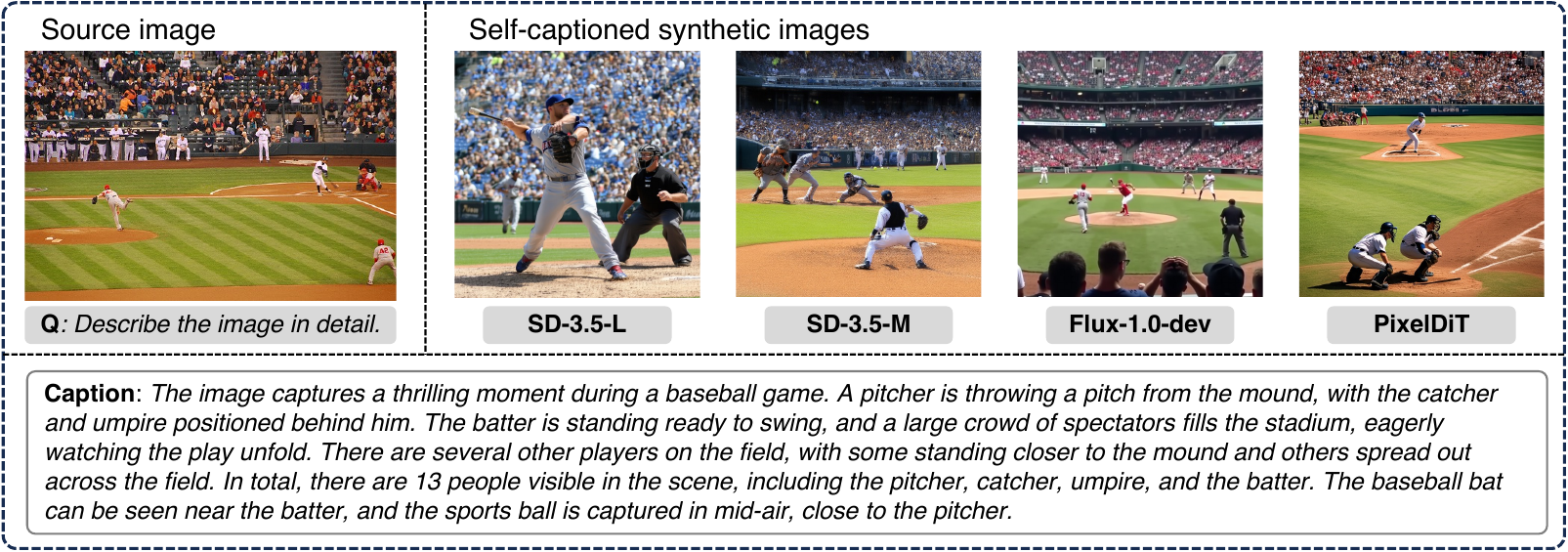}

  \includegraphics[width=\linewidth]{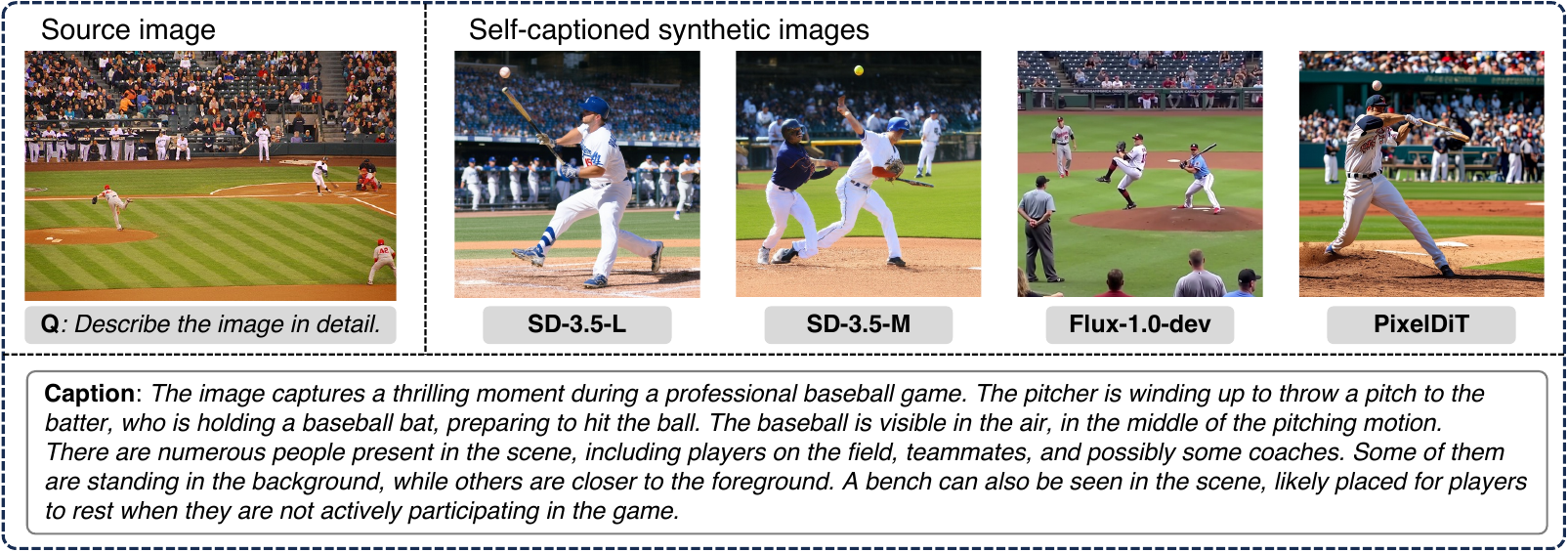}

    \caption{Examples of self-captioned semantic synthesis. The source image is first captioned by the policy model, and the caption is used to generate semantically aligned synthetic images using the text-to-image models SD-3.5-L, SD-3.5-M, Flux-1.0-dev, and PixelDiT.}
  \label{ap:self-cap-synthesis_qual_ex}
\end{figure*}

\begin{figure*}[ht]
  \centering

  \includegraphics[width=\linewidth]{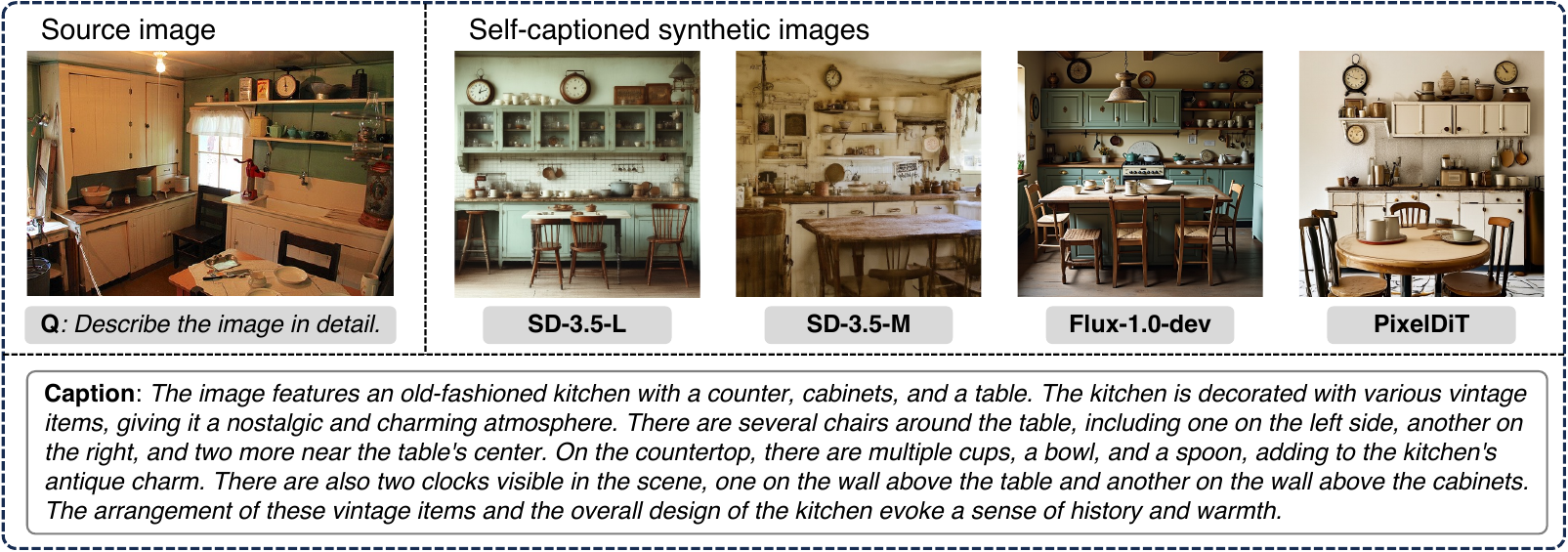}

  \includegraphics[width=\linewidth]{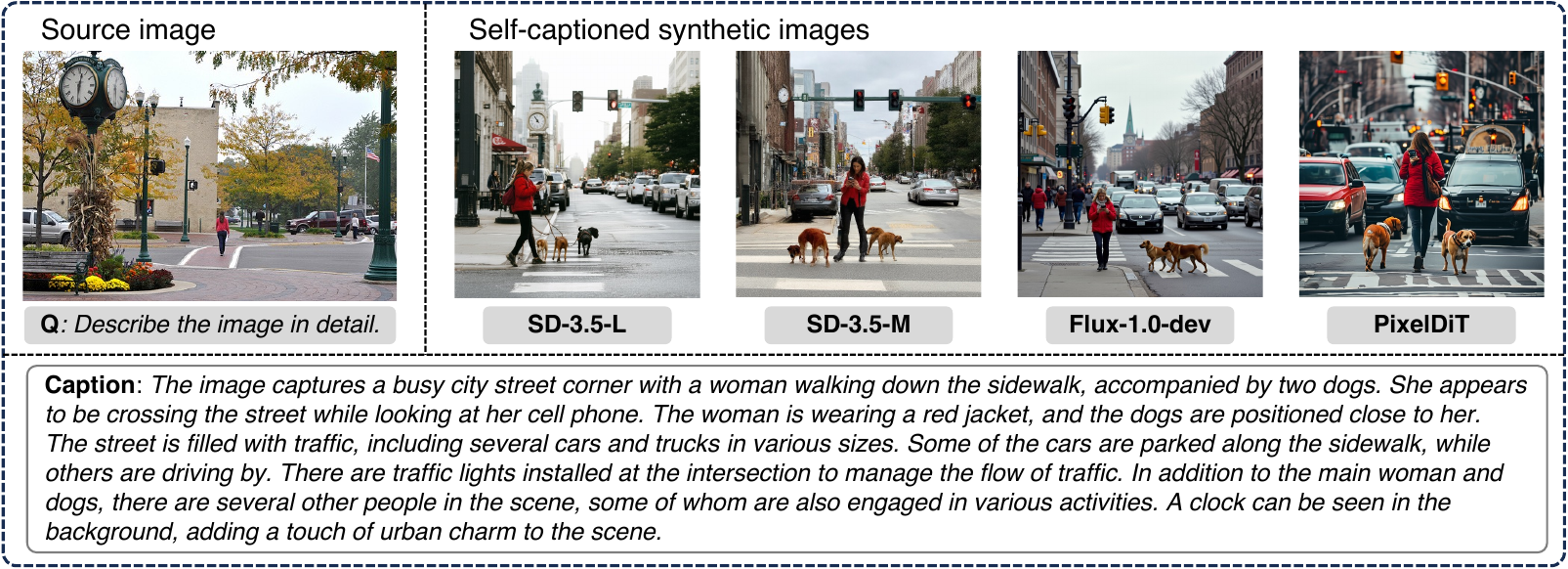}

  \includegraphics[width=\linewidth]{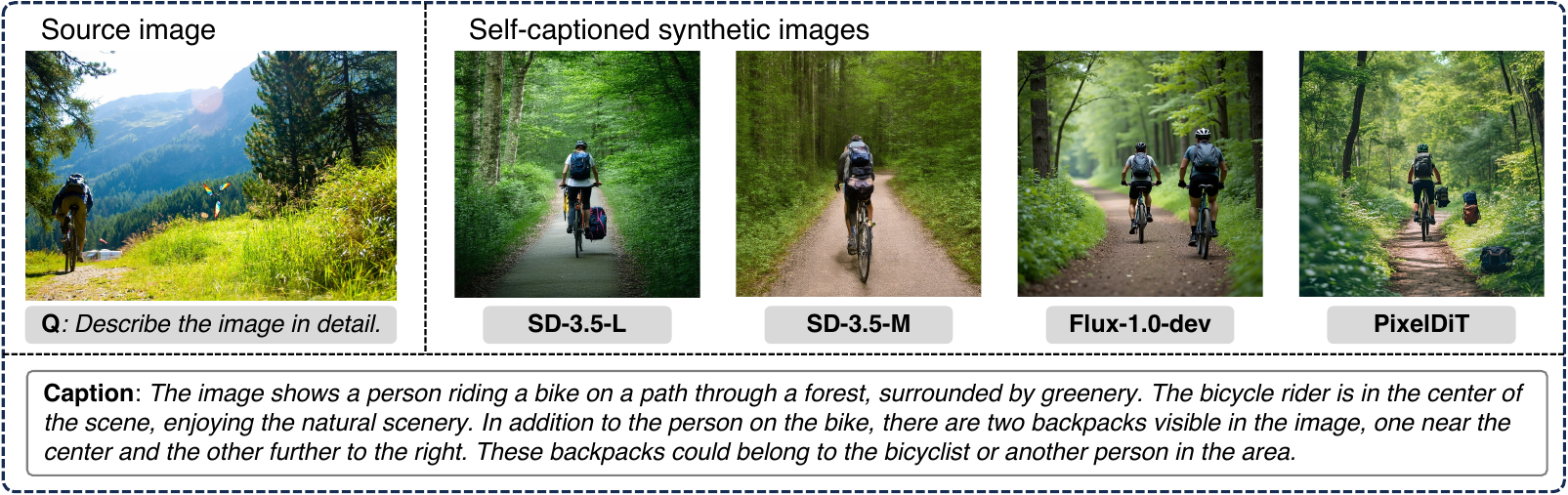}
  
  \includegraphics[width=\linewidth]{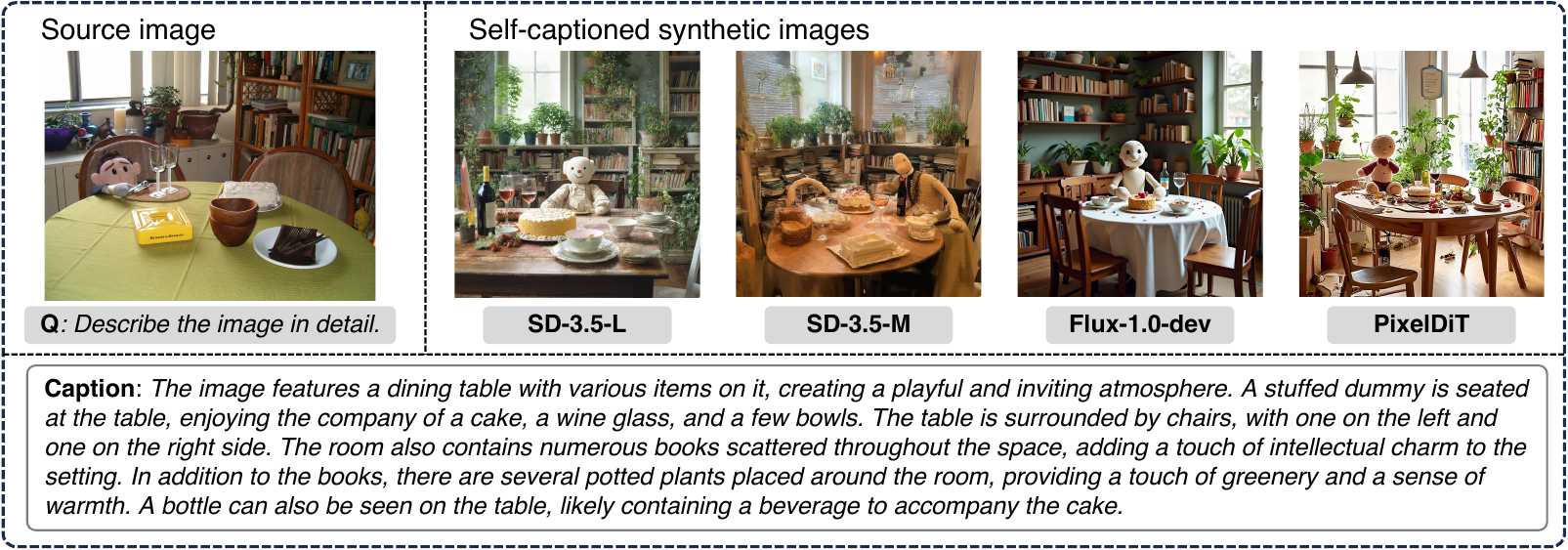}
\caption{Examples of self-captioned semantic synthesis. The source image is first captioned by the policy model, and the caption is used to generate semantically aligned synthetic images using the text-to-image models SD-3.5-L, SD-3.5-M, Flux-1.0-dev, and PixelDiT.}
  \label{fig:ap_selfcap_synthesis_2}
\end{figure*}

Figures~\ref{ap:self-cap-synthesis_qual_ex} and~\ref{fig:ap_selfcap_synthesis_2}
illustrate examples of self-captioned semantic synthesis. For each source image, the policy first generates a self-caption, which is then used as a prompt for text-to-image synthesis. Although the synthetic images vary in appearance, recurring objects and scene elements across the original and synthetic images provide reliable evidence for constructing the visual cue. These examples show how semantic diversity from synthesis helps \methodname{} extract image-grounded object-level cues.

\vspace{-0.05in}
\section{Prompts}
\label{ap:prompt}
\vspace{-0.05in}

\begin{figure}[t]
\centering
\includegraphics[width=0.8\linewidth]{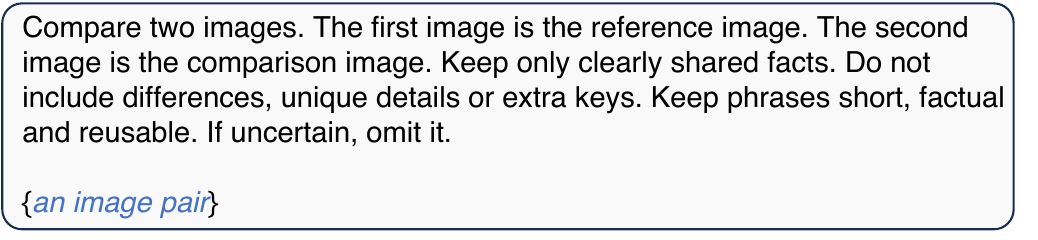}
\caption{Pairwise comparison prompt.}
\label{fig:ap_comparison}
\end{figure}

\begin{figure}[t]
\centering
\includegraphics[width=0.8\linewidth]{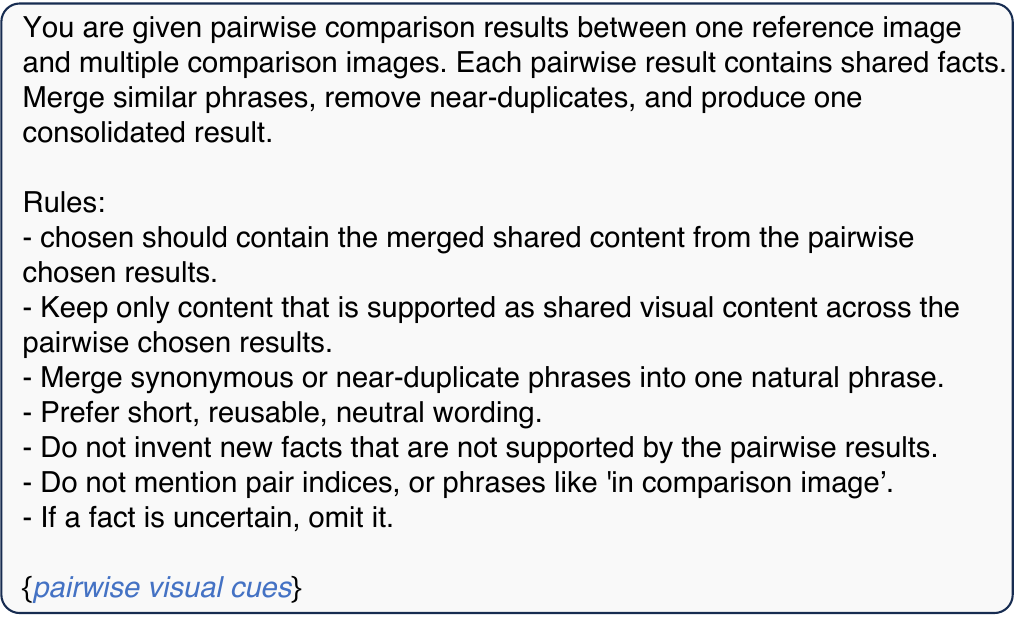}
\caption{Aggregation prompt.}
\label{fig:ap_aggregation}
\end{figure}

\begin{figure}[t]
\centering
\includegraphics[width=0.8\linewidth]{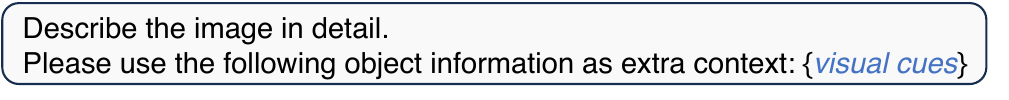}
\caption{Cue-conditioned captioning prompt.}
\label{fig:ap_rollout}
\end{figure}

\begin{figure}[t]
\centering
\includegraphics[width=0.8\linewidth]{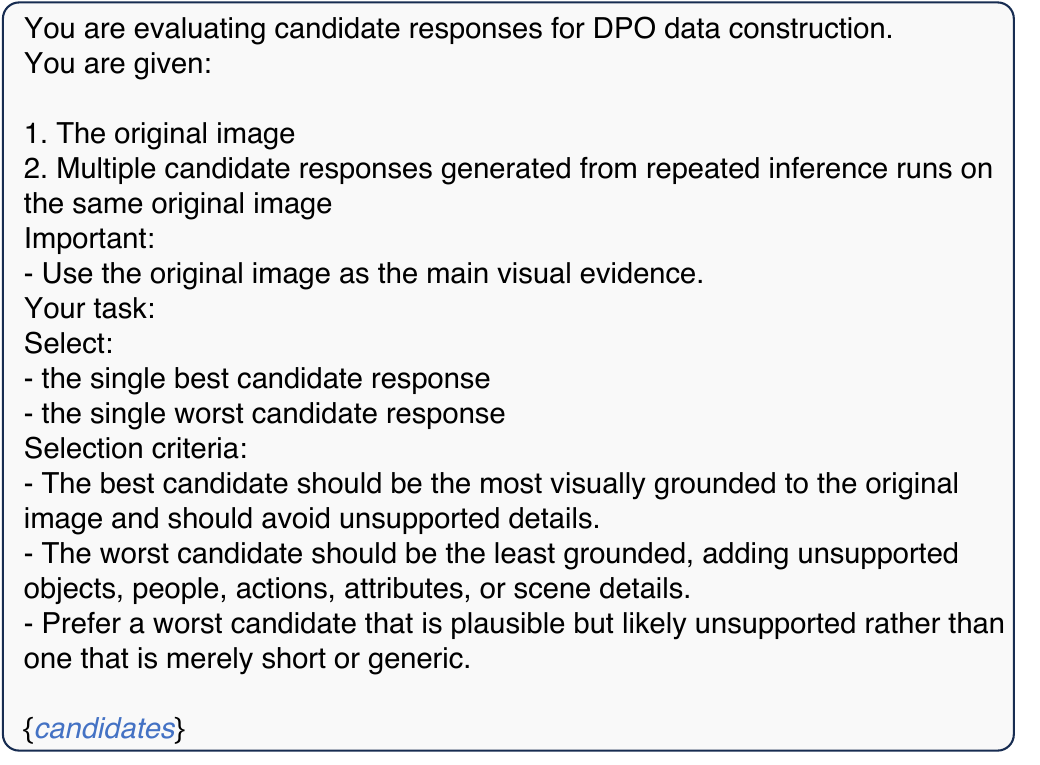}
\caption{Visual grounding prompt.}
\label{fig:ap_visual_grounding}
\end{figure}

We list prompts used in \methodname{} for preference construction.
For self-captioning, we use the prompt ``\textit{Describe the image in detail.}''
The pairwise comparison prompt $\vx_{\text{comp.}}$ in Figure~\ref{fig:ap_comparison},
the aggregation prompt $\vx_{\text{agg.}}$ in Figure~\ref{fig:ap_aggregation},
the cue-conditioned captioning prompt $\vx_{\text{cue-cap.}}$ in Figure~\ref{fig:ap_rollout},
and the visual grounding prompt $\vx_{\text{vis.}}$ in Figure~\ref{fig:ap_visual_grounding}
are used for cue extraction and preference construction.

\clearpage

\end{document}